\documentclass[a4paper,10pt]{IEEEtran}

\usepackage{verbatim} 
\usepackage[table]{xcolor}
\usepackage[utf8]{inputenc}
\usepackage{algorithm,algorithmic}
\usepackage{soul}
\usepackage{empheq}
\usepackage{amsmath,amsthm,amsfonts}
\usepackage{xcolor}
\usepackage{graphicx}
\usepackage{subcaption}
\usepackage{framed}
\usepackage{float}	
\usepackage{multirow}
\usepackage{tabulary}
\usepackage{tcolorbox}
\usepackage[space]{cite}
\usepackage{bm}
\usepackage{authblk}
\usepackage{longtable}

\usepackage{hyperref}
\hypersetup{
  colorlinks   = true,
  urlcolor     = blue,
  linkcolor    = blue,
  citecolor    = blue
}

\theoremstyle{definition}
\newtheorem{defn}{Definition}

\theoremstyle{definition}
\newtheorem{exmp}{Example}

\newcolumntype{L}[1]{>{\raggedright\let\newline\\\arraybackslash\hspace{0pt}}m{#1}}
\newcolumntype{C}[1]{>{\centering\let\newline\\\arraybackslash\hspace{0pt}}m{#1}}
\newcolumntype{R}[1]{>{\raggedleft\let\newline\\\arraybackslash\hspace{0pt}}m{#1}}

\title{%
	Scheduling Plans of Tasks \\ 
	\normalsize Internship report. Supervisors: Amal El Fallah Seghrouchni, Safia-Kedad Sidhoum}

\author[1]{Davide Andrea Guastella}

\date{}

\begin{document}
\maketitle


\begin{abstract}

We present a heuristic algorithm for solving the problem of scheduling plans of tasks. The plans are ordered vectors of tasks, and tasks are basic operations carried out by resources. Plans are tied by temporal, precedence and resource constraints that makes the scheduling problem hard to solve in polynomial time.

The proposed heuristic, that has a polynomial worst-case time complexity, searches for a feasible schedule that maximize the number of plans scheduled, along a fixed time window, with respect to temporal, precedence and resource constraints.
\end{abstract}



\section{Introduction}\label{introduction}

Scheduling is a decision-making process that is used on a regular basis in many manufacturing and services industries~\cite{pinedo_scheduling_2012}. It deals with the allocation of resources to tasks over given time periods and its goal is to optimize one or more objectives.

From a theoretical point of view, a scheduling problem is a constrained combinatorial optimization problem where a set of task must be ordered in such a way that all these are arranged, according to one or more  constraints, to constitute a schedule that minimize or maximize a given objective function.

One of the most popular scheduling problem is the \textit{resource-constrained project scheduling problem} (RCPSP)~\cite{klein_scheduling_2000}: the objective of resource-constrained scheduling consists in developing a schedule such that a set of tasks is completed as early as possible considering both the precedence relationships and the restricted availabilities of resources.

The scheduling problem we are facing can be thought as a particular case of the RCPSP problem: in our case we are facing the problem of scheduling plans of tasks subject to both precedence, resource and temporal constraints.
Also, rather than scheduling tasks as in the RCPSP, our problem aims at selecting a maximum number of plans.

\section{Case studies}\label{Case studies}

Nowadays, airborne platforms such as \textit{Remote Piloted Air Vehicles} (RPAS) are employed in different scenario including conflicts, surveillance and rescue~\cite{grivault_agent-based_2016}. In these scenario, airborne platforms operate in highly dynamic environments with a low predictability. In this context, onboard instruments (i.e. sensors) allow the platform, hence the mission manager, to collect knowledge from the field. Sensors carried by RPAS are now able to perform a large panel of functions such as image acquisition, spectrum analysis, and object tracking~\cite{kemkemian_toward_2010}. All these sensors play a major role in operation and their optimization has become essential.

Because of the criticality of the context and the mission’s objectives, it is important to develop a method that orchestrates the operations conducted by the sensors, such that the mission is accomplished correctly and by satisfying all the constraints. Moreover, due to both critical contexts and dynamic environment, it is important to orchestrate the operations of the sensors within a relatively short time.

As further case study, consider the in-flight airplane safety~\cite{imai_airplane_2017}. In this context there is a need for detecting and resolving data errors which could come from faulty sensor measurements, inaccurate data processing, or poor information transmission, that can lead to catastrophic accidents as in the case of the Air France 447’s accident. Our scheduling technique can be employed as a fault recovery technique: for example, once the fault has been detected by the on-board sensors, a system scheduler could discards the remaining flight plans, and schedules a set of emergency plans into the current flight schedule. These emergency plans are strictly time constrained, and their orchestration necessarily needs to lead to a schedule which can guarantee the safety of passengers.

\section{Definitions}\label{Definitions}

\begin{defn}[Plan of Tasks]
A \textit{plan of tasks} $\Pi_k$ is a partial ordering of tasks $J^k_i$ to address a specific goal. Each plan $\Pi_k$ is characterized by a priority value $\alpha_k$. 
The structure of the plan is depicted by an \textit{activity-on-node} (AON) network where the nodes and the arcs represent the tasks and the precedence relations respectively~\cite{schwindt_handbook_2015}. The precedence relations are described by the notation $\prec$. For example, given two plans $\Pi_k$ and $\Pi_j$, $\Pi_k \prec \Pi_j$ indicates that the plan $\Pi_k$ must be scheduled before $\Pi_j$.

 A plan is defined by the following notation:
	
$$\Pi_k= (J^k_1,...,J^k_{n_k})$$
	
\noindent where $k$ is an arbitrary index for the plan, $n_k$ is the number of tasks in the plan $\Pi_k$ and $J^k_i$, with $0\leq i \leq n_k$ is the $i$-th task of the plan $\Pi_k$. The set of tasks is topologically sorted (see~\ref{def:topological_sorting}).

A plan $\Pi_k$ could not be scheduled if at least one task \mbox{$J^k_i \in \Pi_k$} could not be scheduled due to unsatisfied constraints.

A plan $\Pi_k$ is characterized by a set of resources $\mathcal{R}^k \subseteq \mathcal{R}$.
\end{defn}

\begin{defn}
A \textit{graph} $G=(V,E)$ consists of a set of vertices $V$, and a set of edges $E \subseteq V \times V$. In a directed graph the edges are directed from one vertex to another. A \textit{directed acyclic graph} (or DAG) is a directed graph with no directed cycles: a directed cycle is a path that starts from any vertex $u$ and ends in $u$.
\end{defn}

\begin{defn}[Topological Sort]
\label{def:topological_sorting}
A \textit{topological sort} of a directed acyclic graph (DAG) $G=(V,E)$ is a linear ordering of all its vertices such that if the graph $G$ contains an edge $(u,v)$ then $u$ appears before $v$ in the ordering~\cite{cormen_introduction_2001,kahn_topological_1962}. If the graph contains a cycle, then no linear ordering is possible.
\end{defn}

\begin{figure}[!h]
	\centering
	\includegraphics[scale=0.7]{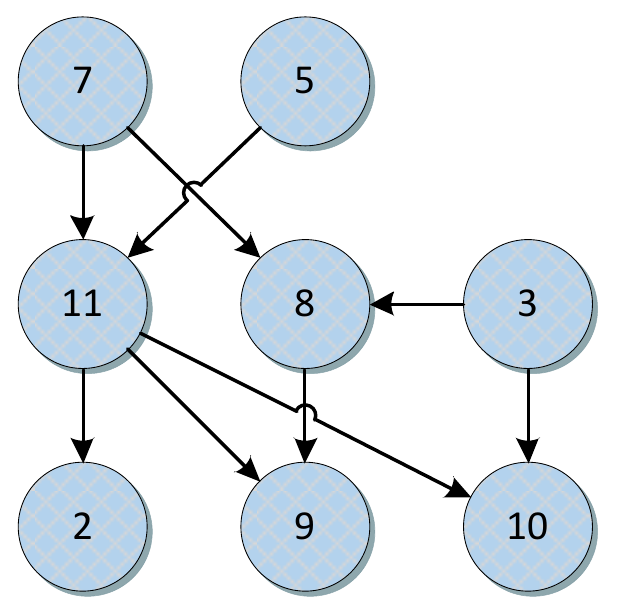}
	\caption{A Directed Acyclic Graph.}
	\label{fig:dag_example}
\end{figure}

\begin{figure}[!h]
	\centering
	\includegraphics[scale=0.7]{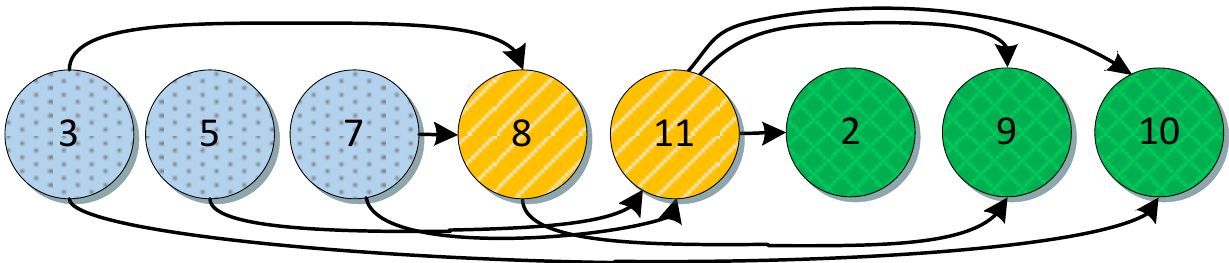}
	\caption{A possible topological sorting for the DAG in Figure~\ref{fig:dag_example}.}
	\label{fig:dag_topological_sorting_example}
\end{figure}

For example, in Figure~\ref{fig:dag_example} a directed acyclic graph is showed, for which a topological sorting can be found, since it has no cycles. Figure~\ref{fig:dag_topological_sorting_example} shows a possible topological sorting for the graph in Figure~\ref{fig:dag_example}, where the different colors depict the different frontiers in the corresponding DFS graph.


\begin{defn}[Frontier]
A \textit{frontier} $f$ for a graph $G=(V,E)$ is a set of nodes $f \subseteq V$ such that the maximum distance between each node $u \in f$ and the root node in the corresponding DFS graph is the same.
\end{defn}

\begin{defn}[Task]A \textit{task} $J^k_i$ is an atomic operation that a subset of resources must execute. Each task $J^k_i$ belongs uniquely to one plan $\Pi_k$. Tasks are characterized by the following elements:
	
	\begin{itemize}
		\item a \textit{processing time} $p^k_i \in \mathbb{N}_{0}$, $p^k_i \geq 1$; 
		\item a \textit{release date} $r^k_i \in \mathbb{N}$;
		\item a \textit{due date} $d^k_i \in \mathbb{N}_{0}$;
		\item a \textit{time lag} $\delta^k_{ij}\in \mathbb{N}$, if $J^k_i \prec J^k_j$;
		\item a \textit{set of resources} $\mathcal{R}^k_i=\{\rho_1,...,\rho_n\}$,  $\mathcal{R}^k_i \subseteq \mathcal{R}^k$, which the task $J^k_i$ is assigned to.
	\end{itemize}
	
	The tasks within a plan can be dependent by a precedence graph that provides the precedence constraints between the tasks. As for the plans, the precedence relations between tasks are described by the notation $\prec$. For example, given two tasks $J_1^k$ and $J_2^k$ belonging to the same plan $\Pi_k$, $J_1^k \prec J_2^k$ indicates that the task $J_1^k$ precedes the execution of the task $J_2^k$.

\begin{defn}[Time lag]	
A \textit{time lag} $\delta^k_{ij}$ is a fixed amount of time which separates the finishing and starting times of a pair of tasks $J^k_i$ and $J^k_j$ respectively. By default, it is $0$.
\end{defn}	
	
\end{defn}

In this paper, we assume that there is no preemption in the task execution. Therefore, the execution of the tasks can not be interrupted once started. Moreover, we assume that each task could consume more than one resource during its execution.

\begin{defn}[Resource]
A \textit{resource} $\rho$ is any hardware or software tool that tasks can use to handle information. A resource $\rho$ has a limited availability value $B_{\rho}$. Resources are typically distinct in renewable and nonrenewable~\cite{schwindt_handbook_2015}: renewable resources have a fixed value of availability in each time period, while nonrenewable resources have a fixed value of availability along the entire project's time horizon. In the addressed scheduling problem, we deal with renewable resources with a fixed availability value of $B_{\rho} = 1,~\forall \rho \in \mathcal{R}$, in each time period. In particular, resources with a fixed availability value of $B_{\rho} = 1$ are also called unary or disjunctive resources.
\end{defn} 

\begin{defn}[Resource availability]
The availability $B_{\rho} \in \mathbb{N}_0$ of a resource $\rho$ represents the maximum value of availability of the resource $\rho$ in each time period.
\end{defn}

The abstract amount of usage of the resource $\rho$ by the task $J^k_i$, in each time period, is represented by $b^k_{i\rho} = \{0,1\}$. 

In this work we are concerned with solving the problem of scheduling plans of tasks by using a heuristic that maximizes the number of plans scheduled in a fixed time window, taking into account precedence, time, and resource constraints.

\section{Problem statement}\label{Problem statement}

\subsection{Constraints}

\subsubsection{Temporal constraints} \label{Temporal constraints}
When a plan $ \Pi_k $ is scheduled, for each task $J^k_i \in \Pi_k$ both the starting time $s^k_i$ and the completion time $C^k_i = s^k_i + p^k_i$ must be inside the temporal window $[W_s,W_e]$ and also inside the temporal window $[r^k_i,d^k_i]$, where 

$$(r^k_i \geq W_s) \wedge (d^k_i \leq W_e),~ \forall J^k_i \in \Pi_k.$$

The value of $s^k_i$ for a task $J^k_i \in Pi_k$ is calculated as the maximum value between $W_s$, the release date $r^k_i$ of the task $J^k_i$ augmented by the time lag $\delta^k_{ij}$, and the maximum completion time $C^k_j$ for each predecessor $J^k_j \in \Pi_k$. Formally, 

\begin{centering}
\begin{equation}\label{eq:initial_starting_time}
\begin{split}
s^k_i = max(W_s, r^k_i, ~max_{j\in pred^k_i}(C^k_j + \delta^k_{ij})),
\end{split}
\end{equation}  
\end{centering}

\noindent where $pred^k_i$ is the set of predecessors of the task $J^k_i$.

Given a set $\mathcal{P}$ of plans, the plan $\Pi_k \in \mathcal{P}$ satisfies the temporal constraints if

$$\exists s^k_i~\forall J^k_i \in \Pi_k : (s^k_i \geq r^k_i) \wedge (s^k_i + p^k_i \leq d^k_i). $$ 

\subsubsection{Precedence constraints}

In order to maintain the precedence constraints between plans, a set $\mathcal{P}$ of plans to be scheduled is topologically ordered, since the precedence between plans can be represented as an acyclic directed graph (DAG), as showed in Figure~\ref{fig:dag_example}.
In our context, a graph $G$ is a tuple $(V,E)$ where $V$ is a set of vertices and $v_k \in V$ represents a plan $\Pi_k \in \mathcal{P}$, and an arc $(u,v)$ represents a precedence constraint between the plans that are represented by nodes $u$ and $v$ respectively.

\subsubsection{Resource constraints}


The scheduling problem we address assumes that each task $J^k_i \in \Pi_k$ which has a feasible starting time $s^k_i$, can be executed by a subset $\mathcal{R}^k_i \subseteq \mathcal{R}^k$ of resources if in the time period $[s^k_i, C^k_i]$ the utilization of each resource $\rho \in \mathcal{R}^k_i$ does not exceed its availability $B_{\rho}$.

\subsection{Classification and mathematical formulation}

A typical formal way to describe the scheduling problems is by using the three-field classification $[\alpha,\beta,\gamma]$ introduced by Graham et al.~\cite{graham_optimization_1979}, where

\begin{itemize}
	\item $\alpha$ specifies the machine environment,
	\item $\beta$ specifies the characteristics of the activities,
	\item $\gamma$ and describes the objective function(s).
\end{itemize}

An extension of the Graham's classification has been proposed in order to provide a more accurate way to formally describe the scheduling problems~\cite{brucker_resource-constrained_1999}. According to the extended notation, the plans of tasks scheduling problem can be stated as

$$ PS1,1,1~|~prec_{\Pi};~prec_{J};~[W_s,W_e];~[r^k_i,d^k_i];~\delta^k_{ij}~|~\max_{k}\sum_{k} \alpha_k x_{k}$$ 

\noindent where $PSm,1,1 \in \alpha$ indicates a resource environment for a project scheduling problem with $m$ resources, a maximum availability of $1$ unit per time for each resource, and a resource utilization, by each task, of at most $1$ unit per time period. $prec_{\Pi};~prec_{J};~[W_s,W_e];~[r^k_i,d^k_i];~\delta^k_{ij} \in \beta$ depicts the characteristics of both plans and tasks: $prec_{\Pi}$ indicates a precedence constraints between plans; $prec_{J}$ indicates a precedence constraints between tasks; $[W_s,W_e]$ and $[r^k_i,d^k_i]$ indicates, for each task $J^k_i \in \Pi_k$, the time constraints discussed in~\ref{Temporal constraints}; $\delta^k_{ij}$ indicates that each task $J^k_i$ can have a time lag between its starting time $s^k_i$ and the completion time $C^k_j$ of its predecessor.

The mathematical formulations for the RCPSP can be conveniently employed to model the objective function with respect to the temporal, precedence and resource constraints~\cite{artigues_resource-constrained_2007}.
The proposed formulation is based on time discretization for describing the usage of the resources and the processing of the tasks over time. In our context the time horizon, divided into unitary time periods, is a fixed time window $[W_s,W_e]$, where the maximum number of plans has to be scheduled. Given a plan $\Pi_k \in \mathcal{P}$ and a task $J^k_i \in \Pi_k$, and a time instant $t \in [r^k_i,d^k_i]$, let $y^k_{it}$ be a boolean variable that indicates whether the task $J^k_i$ starts exactly at time $t$. Also, let $x_k$ be a boolean variable that indicates whether a plan $\Pi_k \in \mathcal{P}$ is executed. Thus, a plan $\Pi_k \in \mathcal{P}$ is executed if

$$\sum_{i=1}^{n_k} \sum_{t=r^k_i}^{d^k_i} y^k_{it} = n_k~\Rightarrow~x_k=1,$$

\noindent where $n_k$ is the number of tasks in $\Pi_k$. If a task $J^k_i$ has a feasible starting time $s^k_i$ equals to $t$, then $y^k_{it} = 1$ for a time instant $t \in [r^k_i,d^k_i]$. Thus, if the sum 

$$\sum_{i=1}^{n_k} \sum_{t=r^k_i}^{d^k_i} y^k_{it}$$ 

\noindent is equal to $n_k$, then $\Pi_k$ is executed.
Therefore, the variable $x_k$ is equal to $1$. 
By using this notation, the objective function to maximize can be formulated as: 

\begin{align} 
\max_{k}\sum_{k=1}^K \alpha_k x_{k}\nonumber
\end{align}

\noindent such that,

\begin{align}
\sum_{i=1}^{n_k} \sum_{t =r^k_i}^{d^k_i} y^k_{it} = n_k \Rightarrow x_k=1 &&\forall \Pi_k \in \mathcal{P} \label{eq:scheduled_plan_condition} \\
(t + \delta^k_{ij} \geq r^k_i)~\wedge \nonumber \\  (t + \delta^k_{ij} + p^k_i \leq d^k_i) \Rightarrow y^k_{it} = 1   && \forall t \in [W_s,W_e], \nonumber \\ &&  \forall J^k_i \in \Pi_k \label{eq:task_in_good_boundaries} \\
\sum_t t \cdot (y^k_{it}-y^k_{jt}) \geq p^k_i + \delta^k_{ij} && \textnormal{if } J^k_i \prec J^k_j, \nonumber \\&&\forall J^k_i, J^k_j \in \Pi_k, i \neq j \label{eq:precedence_constraints}
\end{align}
\begin{align}
\sum_{i=1}^{n_k} b^k_{i\rho} \cdot \sum_{\tau = t}^{t+p^k_i} y_{i\tau}^k \leq B_k && \forall \rho \in \mathcal{R},~\forall t \in [W_s,W_e] \label{eq:resource_constraints}
\end{align}

Constraint~\eqref{eq:scheduled_plan_condition} imposes that for each plan $\Pi_k \in \mathcal{P}$ all the tasks $J^k_i \in \Pi_k$ have a starting time assigned.
Constraint~\eqref{eq:task_in_good_boundaries} imposes that each task $J^k_i$ has a starting time and a completion time within the time window $[r^k_i,d^k_i]$.
Constraints~\eqref{eq:precedence_constraints} and~\eqref{eq:resource_constraints} impose respectively the precedence and resource constraints.

\section{Proposed method}\label{Proposed method}

In this section is discussed a greedy heuristic for the problem of scheduling plans of tasks. 

\begin{defn}[Schedule]
A \textit{schedule} $S$ is a vector $(s^k_i)$, \mbox{$i=1,...,n_k,~k=1,...,K$} such that $s^k_i \in S$ is a starting time for the task $J^k_i$. A schedule represents a solution for the scheduling problem if it satisfies the scheduling constraints.
\end{defn}

Given a temporal window $W = [W_s,W_e]$, the goal of the proposed heuristic is to create a feasible schedule that contains the maximum number of plans of tasks, inside the temporal window $[W_s,W_e]$, regarding their weight, such that each scheduled plan $\Pi_k$ satisfies the precedence, temporal and resource constraints discussed in~\ref{Problem statement}. The set $\mathcal{P}$ of plans given as input to the algorithm is preventively sorted in order to maintain their precedence relations, and also to evaluate, at each iteration of the algorithm, the plan with the highest priority value. The feasibility of the plans is evaluated in a schedule $S_w$ until no plans remains to schedule, or there is no plan that can be scheduled due to a violation of the constraints.

Each plan $\Pi_k \in \mathcal{P}$ could have precedence relations between its tasks. The topological sorting of the tasks ensures that the execution of the tasks is done with respect to the their precedence relations. Example~\ref{example_1} shows an example of feasible schedule where a precedence relation occurs between the tasks of a plan.

\begin{exmp} 
\label{example_1}
Figure~\ref{fig:wrong_precedence} shows a Gantt diagram for a feasible schedule for a set of plans $\mathcal{P}=\{\Pi_1,\Pi_2\}$, which parameters are listed in Table~\ref{table:example_1_table}. A precedence constraint occurs between the tasks $J^2_1$ and $J^2_2$, both belonging to the plan $\Pi_2$.

\begin{table}[!h]
\centering
\begin{tabular}{lllllll}
\hline
        & $r^k_i$ & $d^k_i$ & $p^k_i$  & $\delta^k_{ij}$  &$\mathcal{R}^k_i$ & $J^k_i \prec J^k_j$\\ \hline
$J^1_1$ & $2$     & $7$  & $3$  & $0$ &$\{1\}$ & \\ \hline
$J^2_1$ & $3$     & $8$  & $2$  & $0$ &$\{2\}$ & $J^2_1 \prec J^2_2$\\ \hline
$J^2_2$ & $4$     & $9$  & $2$  & $0$ &$\{1\}$ &\\ \hline
\end{tabular}
\caption{The parameters of the tasks of the plans $\Pi_1$ and $\Pi_2$ in the Example~\ref{example_1}.}
\label{table:example_1_table}
\end{table}

\begin{figure}[!ht]
	\centering
	\includegraphics[scale=1.1]{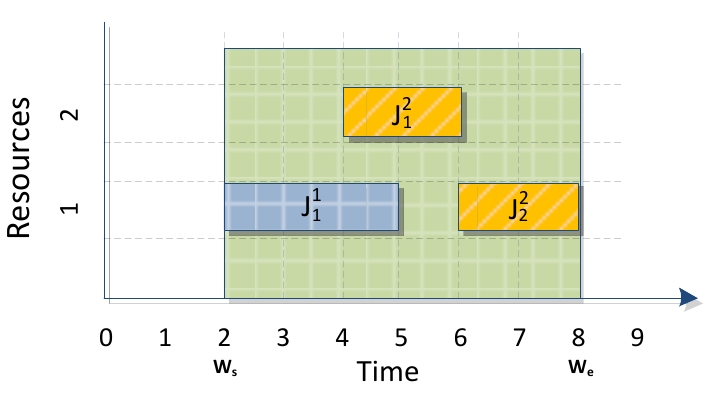}
	\caption{A feasible schedule for the set $\mathcal{P}$ of two plans. A precedence constraint occurs between $J^2_1$ and $J^2_2$.}
	\label{fig:wrong_precedence}
\end{figure}

The task $J^2_2$ has been scheduled after the end of the execution of the task $J^2_1$, since there is a precedence constraint between $J^2_1$ and $J^2_2$. For this reason, the task $J^2_2$ has not been scheduled at $5$.

\end{exmp}


For a given DAG, more than one topological sort could exist. Under this assumption, different topological sorting for a plans' precedence graph could lead to different schedules. For this reason, the plans belonging to the same frontier are sorted by their priority value, since for these plans the order in which they are scheduled is irrelevant according to the precedence constraints. Moreover, different plans may have the same priority value: in this case, the plans must be sorted according to a different criterion described in the following.

Given a feasible schedule $S_w$ and a subset $\mathcal{P}_{\alpha_k} \subseteq \mathcal{P}$ of plans having the same priority value $\alpha_k$, in order to maximize the resource utilization for each tasks $J^k_i \in \Pi_k \in \mathcal{P}_{\alpha_k}$ we calculate an idle time between each task $J^k_i \in \Pi_k$ and its predecessor, in the same resource. 

\begin{defn}[Idle time]
Let $S$ be a feasible schedule containing two plans $\Pi_k$, $\Pi_p$, $p \neq k$. Let $J^k_i \in \Pi_k$ and $J^p_j \in \Pi_p$ be two tasks scheduled on the same resource $\rho \in \mathcal{R}$, where $s^k_i > C^p_j$. If there is at least one plan $\Pi_q$ schedulable in $S$ such that $\exists J^q_l \in \Pi_q$ executed by the same resource $\rho \in \mathcal{R}$ such that 

$$(C^p_j \leq s^q_l < s^k_i) \wedge (C^p_j < C^q_l \leq s^k_i),$$

\noindent the time window $[C^p_j, s^k_i]$ is said to be an \textit{idle time}.

\end{defn}

\begin{figure}[!h]
	\centering
	\includegraphics[scale=0.8]{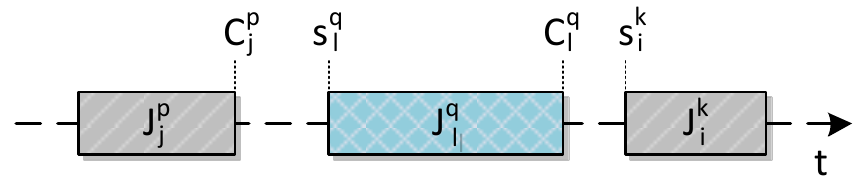}
	\caption{The time window $[C^p_j, s^k_i]$ is an idle time because \mbox{$J^q_l \in \Pi_q$} can be scheduled in this interval.}
	\label{fig:idle_time_definition}
\end{figure}

In Figure~\ref{fig:idle_time_definition} is showed an idle time between two tasks.

Let $S$ be a feasible schedule, and $\mathcal{P}_{\alpha_k}$ a set of plans having each one priority $\alpha_k$. The plans of $\mathcal{P}_{\alpha_k}$ are scheduled in order to minimize the presence of idle times. This is done by choosing iteratively the plans who minimize the following quantity:

$$\sum_{i} s^k_i - C_{\Gamma^k_i(\rho_i)},~\forall J^k_i \in \Pi_k \in \mathcal{P}_{\alpha_k}$$

\noindent where $C_{\Gamma^k_i(\rho_i)}$ is the completion time of the task $\Gamma^k_i(\rho_i)$, that is, the predecessor of task $J^k_i$ in the same resource $\rho_i$ executing the task $J^k_i$.

Example~\ref{example_idle_time} shows how the idle time is used by the proposed heuristic to determine the order according to which two plans have to be scheduled.

\begin{exmp}
\label{example_idle_time}

Let $\mathcal{P}=\{\Pi_1,\Pi_2,\Pi_3,\Pi_4\}$ be a set of plans to schedule, which information are listed in Table~\ref{table:plans_with_same_priority_data}, and let $S_w$ be a feasible schedule where $\Pi_1$ and $\Pi_2$ have been scheduled (Figure~\ref{fig:dead_times_plan01}). Let $\mathcal{P}_l = \{ \Pi_3, \Pi_4 \}$ be a set of two plans such that $\Pi_3$ and $\Pi_4$ have priority values $\alpha_3 = l$ and $\alpha_4 = l$ respectively. As showed in Figure~\ref{fig:dead_times_plan02}-\ref{fig:dead_times_plan03}, both plans $\Pi_3$ and $\Pi_4$ generate two feasible schedules. 

\begin{table}[]
\centering
\begin{tabular}{lllllll}
\hline
    & $r^k_i$ & $d^k_i$ & $p^k_i$  & $\delta^k_{ij}$  & $\mathcal{R}^k_i$ & $J^k_i \prec J^k_j$ \\ \hline
$J^1_1$ & $2$  & $7$  & $3$   & $0$ & $\{1\}$ & \\ \hline
$J^2_1$ & $2$  & $6$  & $2$  & $0$ &$\{2\}$ &\\ \hline
$J^2_2$ & $4$  & $10$  & $3$  & $0$ &$\{1\}$ &\\ \hline    
$J^3_1$ & $4$  & $7$  & $2$  & $0$ &$\{3\}$ &\\ \hline
$J^4_1$ & $3$  & $6$  & $1$  & $0$ &$\{4\}$ & $J^4_1 \prec J^4_2$\\ \hline
$J^4_2$ & $2$  & $7$  & $3$  & $0$ &$\{2\}$ &\\ \hline
\end{tabular}
\caption{The parameters of the tasks of a set $\mathcal{P}$ of $4$ plans.}
\label{table:plans_with_same_priority_data}
\end{table}

The difference between the starting time $s^3_1$ and its predecessor (in this case $W_s$ since there is no predecessor of $J^3_1$) is equal to 2. 
Within $\mathcal{P}_l$, the plan $\Pi_4$ has a task $J^4_1$ that can be scheduled at $s^4_1 = 3 < s^3_1$, where all the constraints are satisfied, and both $J^4_1$ and $J^3_1$ demand the same resource.
Moreover, scheduling the plan $\Pi_4$ before $\Pi_3$ leads to a schedule where no idle times are generated. Therefore, plan $\Pi_4$ is scheduled before plan $\Pi_3$.

\begin{figure}[!ht]
	\centering
	\begin{subfigure}{\linewidth}
		\centering
		\includegraphics{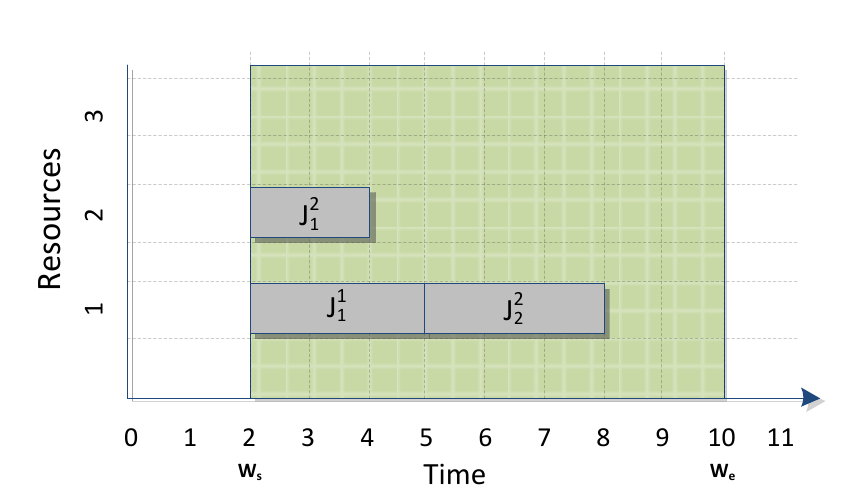}
		\caption{}
		\label{fig:dead_times_plan01}
	\end{subfigure}%
	
	\begin{subfigure}{\linewidth}
		\centering
		\includegraphics{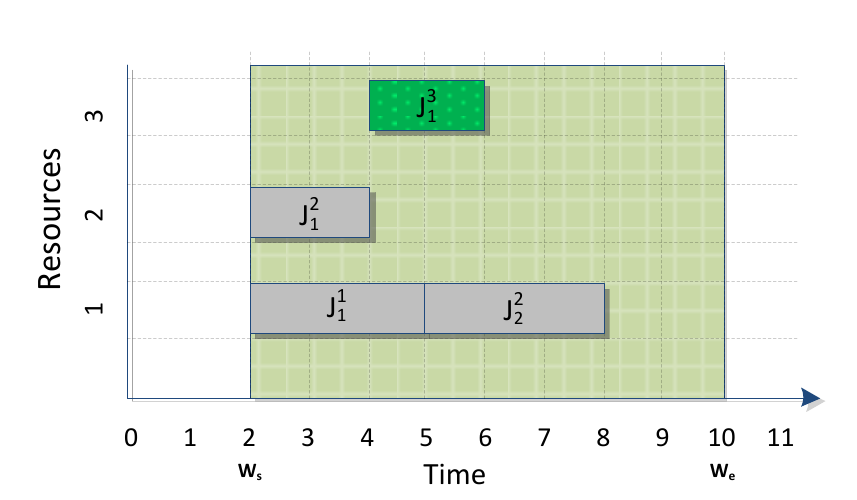}
		\caption{}
		\label{fig:dead_times_plan02}
	\end{subfigure}
	
	\begin{subfigure}{\linewidth}
		\centering
		\includegraphics{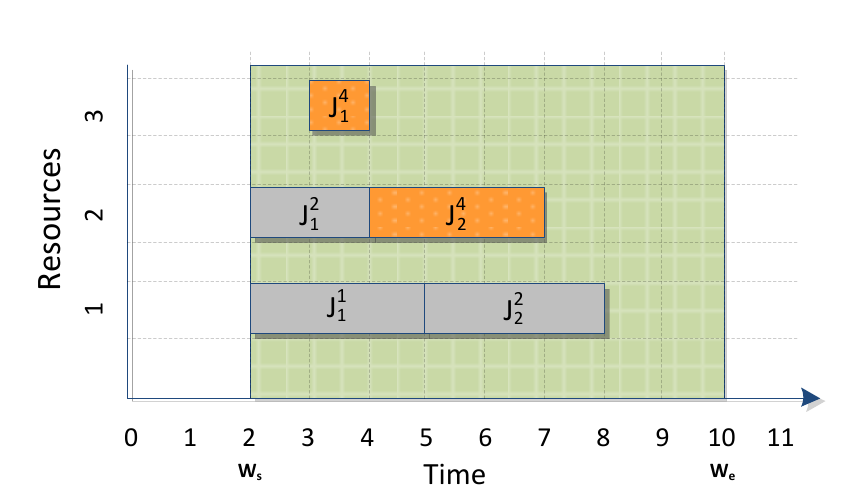}
		\caption{}
		\label{fig:dead_times_plan03}
	\end{subfigure}
	\caption{Scheduling plans with the same priority value into a schedule with two plans schedule (Figure~\ref{fig:dead_times_plan01}). Scheduling plan $\Pi_3$ before plan $\Pi_4$ (Figure~\ref{fig:dead_times_plan02}) leads to a schedule with an idle time of 2, since in the interval $[2,4]$ the task $J^4_1$ of the feasible plan $\Pi_4$ can be scheduled. Scheduling plan $\Pi_4$ before $\Pi_3$ (Figure~\ref{fig:dead_times_plan03}) leads to a schedule with an idle time of 0. Therefore, $\Pi_3$ is scheduled before $\Pi_4$.} 
	\label{fig:dead_times_example}
\end{figure}

\end{exmp}

The scheduling algorithm operates in a similar way as the serial scheduling scheme proposed in~\cite{artigues_resource-constrained_2007}. In our context, we assume that each resource has a maximum availability of 1 in each time period. In this case, each resource can execute one task at a time. The idea of the proposed heuristic is to schedule all the plans as soon as possible by respecting the resource, precedence and temporal constraints. The insertion of the plans into a schedule is done by handling an \textit{event list} $\mathcal{EL}$. An \textit{event} is a $4$-tuple $(t(e),\mathcal{S}(e),\mathcal{C}(e),(b_{\rho}(e))_{\rho\in \mathcal{R}})$:

\begin{itemize}
	\item $t(e)$ denotes the time instant of the event;
	\item $\mathcal{S}(e)$ is the set of tasks starting exactly at time $t(e)$;
	\item $\mathcal{C}(e)$ is the set of tasks completing exactly at time $t(e)$;
	\item $b_{\rho}(e)$ represents the usage of resource $\rho$ during interval $[t(e),t(next_e)[$, where $next_e$ is the first event following $e$ in $\mathcal{EL}$.
\end{itemize}
		
When a feasible task $J^k_i$ is scheduled and inserted into $\mathcal{S}(e)$ for a an event $e \in \mathcal{EL}$, the usage of the resource $b_\rho(e)$ is set to $1$.
The resource constraints for the scheduling problem we address assumes that no task can be scheduled at $t(e)$ if $b_{\rho}(e) = 1$. 
		
		\subsection{Description of the heuristic}\label{Description of the algorithm}
		
		The proposed heuristic (Algorithm~\ref{alg:scheduleMain}) ($BuildSchedule$) takes in input a set $\mathcal{P}$ of plans and a set $\mathcal{R}$ of resources, and it gives as output a feasible schedule $S_l$, that is, a vector of starting times $s^k_i$ for each scheduled task $J^k_i \in \Pi_k$ such that the constraints discussed in the previous section are met. 
		
		At the beginning, the algorithm initializes the used data structures. At line~\ref{alg_desc:xs_init} and ~\ref{alg_desc:xf_init} respectively, two set $\mathcal{P}_s$ and $\mathcal{P}_f$ are initialized: they will be used to store respectively the successfully scheduled plans and the discarded (unscheduled) plans. Then, two empty schedules $S_l$ and $S_w$ are initialized. They will contain respectively the last feasible schedule found and the current working schedule. The working schedule is a schedule used for testing the feasibility of the  plans during the search. Finally, an event list $\mathcal{EL}$ is initialized at line~\ref{alg_desc:EL_init}.
		The first step of the algorithm is to sort the set $\mathcal{P}$ of plans according to both precedence relations and priority values (line~\ref{alg_desc:sort_set_of_plans}). This is done by Algorithm~\ref{alg:sortPlans}, that returns a sorted list $\mathcal{P}_u$ of plans.
		Then, a loop over $\mathcal{P}_u$ starts until there is a plan left to schedule (line~\ref{alg_desc:alg_main_loop}). At line~\ref{alg_desc:alg_main_pop_plan}, the highest priority plan $\Pi_k$ is taken from the sorted set $\mathcal{P}_u$ of plans. Then, at line~\ref{alg_desc:alg_main_check_priority} the algorithm checks if there exists more than one plan in $\mathcal{P}_u$ with the same priority value $\alpha_k$ of $\Pi_k$. If the value of $\alpha_k$ is unique among the values of priority of the plans in $\mathcal{P}_u$, the algorithm tries to schedule the plan (line~\ref{alg_desc:alg_main_schedule_single_plan}), otherwise a set $\mathcal{P}_{\alpha_k} \subset \mathcal{P}$ of plans is calculated, containing all the plans that have priority value $\alpha_k$. All the plans in $\mathcal{P}_{\alpha_k}$ are then scheduled by Algorithm~\ref{alg:schedulePlanWithSamePriority}. Thus, the plans successfully scheduled are added into $\mathcal{P}_s$ (line~\ref{alg_desc:alg_main_add_successful_plans_to_solution}), and the plans not scheduled are removed from the working schedule (line~\ref{alg_desc:alg_main_remove_unscheduled_from_solution}). Finally, the last feasible schedule $S_l$ is updated (line~\ref{alg_desc:alg_main_update_feasible_solution}). 

Algorithm~\ref{alg:scheduleMain} takes at most $O(K^3n^3)$ time, where $K=|\mathcal{P}|$ is the number of plans to schedule.

All the data structures used for storing plans, $\mathcal{P}$, $\mathcal{P}_u$, $\mathcal{P}_s$ and $\mathcal{P}_f$, are implemented as self-balancing binary search trees to ensure insertion, removal and get operations in logarithmic time~\cite{cormen_introduction_2001}. At line~\ref{alg_desc:EL_init} the event list $\mathcal{EL}$ is initialized. As the data structures used for storing plans, the event list is implemented as self-balancing binary search tree. At line~\ref{alg_desc:sort_set_of_plans} the plans are sorted by the Algorithm~\ref{alg:sortPlans}.

The while loop~\ref{alg_desc:alg_main_loop}-\ref{alg_desc:alg_main_loop_end} iterates until the set of plans $\mathcal{P}_u$ is not empty. In the worst case, if the plans belongs all to the same frontier, the inner loop takes at most $O(K^3n^3)$ time.

The $Get$ operations at line~\ref{alg_desc:alg_main_pop_plan} takes logarithmic time with respect to the number of elements in $\mathcal{P}_u$.

At line~\ref{alg_desc:alg_main_get_number_of_plan_with_priority_alpha_k} the algorithm get the number of plans in $\mathcal{P}_u$ that have priority value $\alpha_k$. We use an auxiliary data structure to store this information and get the value in $O(1)$.

The $Add$ and $Remove$ operations at line~\ref{alg_desc:alg_main_add_to_Ps} and~\ref{alg_desc:alg_main_remove_from_S_w} respectively require both $O(\log K)$ time. Getting the plans with priority $\alpha_k$ (line~\ref{alg_desc:alg_main_get_plans_with_same_priority}) requires constant time, since a hash map is used to store the plans with the same priority value: the key of a record is a value of priority $\alpha_k$, and the value is a list of plans having each one $\alpha_k$ as priority value.

The $Add$ operation at line~\ref{alg_desc:alg_main_add_successful_plans_to_solution} requires $O(K\log K)$ time. The $Remove$ operations at line~\ref{alg_desc:alg_main_remove_plans_from_input_list} and~\ref{alg_desc:alg_main_remove_unscheduled_from_solution} require both $O(K \log K)$ time.

		
		\begin{algorithm}[!h]
			\caption{BuildSchedule}
			\label{alg:scheduleMain}
			\begin{algorithmic}[1]
				\REQUIRE A set of plans $\mathcal{P}$
				\ENSURE A feasible schedule $S_l$, a set $\mathcal{P}_s$ of scheduled plans, a set $\mathcal{P}_f$ of unscheduled plans
				\STATE $\mathcal{P}_s \leftarrow \{\}$ \COMMENT{Scheduled plans} \label{alg_desc:xs_init}
				\STATE $\mathcal{P}_f \leftarrow \{\}$ \COMMENT{Discarded plans} \label{alg_desc:xf_init}
				\STATE $S_l \leftarrow \left(\right)$ \COMMENT{Feasible schedule}  \label{alg_desc:Sl_init}
				\STATE $S_w \leftarrow \left(\right)$ \COMMENT{Working schedule} \label{alg_desc:Sw_init}
				\STATE $\mathcal{EL} \leftarrow \{\}$ \label{alg_desc:EL_init}
				\STATE $\mathcal{P}_u \leftarrow$ SortPlans($\mathcal{P}$) \label{alg_desc:sort_set_of_plans}
				\WHILE{$\mathcal{P}_u \neq \emptyset$}  \label{alg_desc:alg_main_loop}
				\STATE $\Pi_k \leftarrow$ Get($\mathcal{P}_u$)  \label{alg_desc:alg_main_pop_plan}
				
				\STATE $\mathcal{N}_{\alpha_k} \leftarrow$ number of plans in $\mathcal{P}_u$ with priority $\alpha_k$ \label{alg_desc:alg_main_get_number_of_plan_with_priority_alpha_k}
				
				\IF{$\mathcal{N}_{\alpha_k} = 1$} \label{alg_desc:alg_main_check_priority}
				\STATE $success \leftarrow$ SchedulePlan($\Pi_k$, $S_w$, $\mathcal{EL}$)  \label{alg_desc:alg_main_schedule_single_plan}
				\IF{ $success$ }  \label{alg_desc:alg_main_check_plan_scheduled}
				\STATE Add($\Pi_k$, $\mathcal{P}_s$) \label{alg_desc:alg_main_add_to_Ps}
				\ELSE
				\STATE Remove($\Pi_k$, $S_w$) \label{alg_desc:alg_main_remove_from_S_w}
				\ENDIF
				        
				\ELSE
        \STATE $\mathcal{P}_{\alpha_{k}} \leftarrow$ plans of $\mathcal{P}$  with priority $\alpha_{k}$ \label{alg_desc:alg_main_get_plans_with_same_priority}
				\STATE $\mathcal{U} \leftarrow$ SchedulePlanSet($\mathcal{P}_{\alpha_{k}}$, $S_w$, $\mathcal{EL}$) \label{alg_desc:alg_main_schedule_plans_with_same_priority}
				\STATE Add($\mathcal{P}_{\alpha_{k}} \setminus \mathcal{U}$, $\mathcal{P}_s$) \label{alg_desc:alg_main_add_successful_plans_to_solution}
				\STATE Remove($\mathcal{P}_{\alpha_{k}}$, $\mathcal{P}_u$)  \label{alg_desc:alg_main_remove_plans_from_input_list}
				\STATE Remove($\mathcal{U}$, $S_w$) \label{alg_desc:alg_main_remove_unscheduled_from_solution}
				\ENDIF
				      
				\STATE $S_l \leftarrow S_w$ \label{alg_desc:alg_main_update_feasible_solution}
				\STATE $\mathcal{P}_f = \mathcal{P} \setminus \mathcal{P}_s$
				\ENDWHILE \label{alg_desc:alg_main_loop_end}
			\end{algorithmic}
		\end{algorithm}

    Algorithm~\ref{alg:sortPlans} ($SortPlans$) sort a set $\mathcal{P}$ of plans in decreasing order of priority so that the precedence constraints between plans are also respected. At line~\ref{alg:sortPlans_sortTopologically} a topological sort algorithm is executed for the plan set. A set of couples $(\Pi_i,F_i)$ is returned, where $\Pi_i$ is the $i$-th plan in the topological sorting, and $F_i$ is the subset of plans of the frontier $f_i$ which the plan $\Pi_i$ belongs to. Then, let $F$ be the set of subsets $F_i$ of plans for each frontier $f_i$ (line~\ref{alg:set_F_of_frontiers}), a loop over each set $F_i$ is done (line~\ref{alg:sortPlans_loopStart}-\ref{alg:sortPlans_loopEnd}), where each $F_i$ is sorted according to the priority value the plans in the frontier $f_i$ (line~\ref{alg:sortPlans_sortByPriority}), and then added to the set $\mathcal{P}_u$ of sorted plans (line~\ref{alg:sortPlans_addToPu}).


Algorithm~\ref{alg:sortPlans} takes at most $O(K \log K + n_k)$, where $K = |\mathcal{P}|$ is the number of the plans to be sorted. At line~\ref{alg:sortPlans_sortTopologically}, the set of plans is topologically sorted. The topological sort requires at most $O(K+n_k)$ where $n_k$ is the number of predecessors in the precedence's graph of the plan $\Pi_k$.

At line~\ref{alg:set_F_of_frontiers} the set of frontiers discovered by the topological sort is assembled: this set is implemented as an hash table where the key of a record is a priority value $\alpha_k$, and the value is a list of plans having each one $\alpha_k$ as priority value.

The loop~\ref{alg:sortPlans_loopStart}-\ref{alg:sortPlans_loopEnd} iterates over each frontier in the precedence's graph, and sort the plans belonging to that frontier according to their priority value. At line~\ref{alg:sortPlans_sortByPriority}, plans belonging to a specific frontier $f$ are sorted by using the timsort algorithm~\cite{peters_timsort_2002}. The for loop requires at most $O(K\log K)$ time if all the plans belong to the same frontier.

		\begin{algorithm}[!h]
			\caption{SortPlans}
			\label{alg:sortPlans}
			\begin{algorithmic}[1]
				\REQUIRE A set of plans $\mathcal{P}$
				\ENSURE A sorted set of plans $\mathcal{P}_{u}$
				\STATE $\mathcal{P}_{t} \leftarrow TopologicalSort(\mathcal{P})$ \label{alg:sortPlans_sortTopologically}
				\STATE compute $F_i$ for $i=1,...,f$ \label{alg:set_F_of_frontiers}

				\STATE $\mathcal{P}_{u} = \emptyset$
				\FOR{$i=1,...,f$} \label{alg:sortPlans_loopStart}

  				\STATE sort $F_i$ according to $\alpha$ \label{alg:sortPlans_sortByPriority}
				\STATE $Add(F_i, \mathcal{P}_{u})$  \label{alg:sortPlans_addToPu}
				\ENDFOR \label{alg:sortPlans_loopEnd}
				
        \RETURN $\mathcal{P}_{u}$
			\end{algorithmic}
		\end{algorithm}

		Algorithm~\ref{alg:schedulePlan} ($SchedulePlan$) schedules a single plan $\Pi_k$ into a schedule $S_w$ using an event list $\mathcal{EL}$. It iterates over all the tasks $J^k_i \in \Pi_k$ and for each task, the precedence constraints are checked, and then the task is scheduled into the schedule $S_w$. If at least one task could not be scheduled, the algorithm immediately breaks the loop, and false is returned, since the plan could not be scheduled such that all the constraints are satisfied.
		
Algorithm~\ref{alg:schedulePlan} takes $O(n^3)$ time, where $n= \sum_k n_k, \forall \Pi_k \in \mathcal{P}$ is the total number of tasks already scheduled in $S_w$. Inside the loop the Algorithm~\ref{alg:scheduleTask} is executed $O(n_k)$ time.

		\begin{algorithm}[!h]
			\caption{SchedulePlan}
			\label{alg:schedulePlan}
			\begin{algorithmic}[1]
				\REQUIRE A plan $\Pi_k$, a schedule $S_w$, an event list $\mathcal{EL}$
				\ENSURE $true$ if $\Pi_k$ has been scheduled in $S_w$, $false$ otherwise.
				\FORALL{task $J^k_i \in \Pi_k$} 
				\STATE \textit{success} $\leftarrow$ scheduleTask($J^k_i$, $S_w$, $\mathcal{EL}$) \label{alg_desc:alg_schedule_plan_schedule_task}
				\IF{\NOT \textit{success}}
				\STATE mark $\Pi_k$ as unschedulable
				\RETURN $false$
				\ENDIF
				\ENDFOR
				\RETURN $true$
			\end{algorithmic}
		\end{algorithm}

		Algorithm~\ref{alg:scheduleTask} ($ScheduleTask$) do the actual task insertion into a schedule. It starts by calculating the earliest possible starting time given by~\eqref{eq:initial_starting_time} if we do not consider the resource availability. At line~\ref{alg_desc:alg_schedule_task_get_previous_event} the algorithm searches for an event $e$ in $\mathcal{EL}$ such that $t(e) = s^k_i$. If such event does not exist, a new event $e$ is created at $t(e) = s^k_i$. At line~\ref{alg_desc:alg_schedule_task_loop_until_end}, the algorithm starts a loop to search for the earliest feasible start and end events $e$ and $f$ for the task $J^k_i$. The loop iterates until $f$ is not the last event or the remaining duration $\mu$, initially set to $p^k_i$, it is not null. If the condition of the while loop is satisfied, $g$ is assigned the event following $f$ (line~\ref{alg_desc:alg_schedule_task_assign_nextf_to_g}), and the algorithm proceeds by checking the resource availability and the constraints test for $J^k_i$ in $t(e)$ (line~\ref{alg_desc:alg_schedule_task_check_constraints}). If both the tests succeed, $\mu$ is decreased by $t(g)-t(f)$ and $g$ is assigned to $f$. If the tests fail, then $e$ is not a valid insertion position, and then $e$ is set to $g$ while $\mu$ is reset to $p^k_i$. After the loop, a final check for the constraints is done (line~\ref{alg_desc:alg_schedule_task_check_final_constraints}). If the final check fails, then all the tasks of the plan $\Pi_k$ are removed from each event $e$ in the event list $\mathcal{EL}$ (line~\ref{alg_desc:alg_schedule_task_remove_plan_from_events}), the plan $\Pi_k$ is marked as unschedulable (line~\ref{alg_desc:alg_schedule_task_mark_plan_as_unschedulable}) and $false$ is returned.
		
Once the constraints are verified for $e$ at line~\ref{alg_desc:alg_schedule_task_check_final_constraints}, the algorithm proceed with the insertion of the task $J^k_i$ into the working schedule (line~\ref{alg_desc:alg_schedule_task_add_to_solution}), and into the event $e$ (line~\ref{alg_desc:alg_schedule_task_add_to_S}). Then, at line~\ref{alg_desc:alg_schedule_task_add_e_to_EL}, the event $e$ is added into the event list $\mathcal{EL}$ if it is not already in.
The steps (\ref{alg_desc:events_update_start}-\ref{alg_desc:events_update_end}) update the event list. The last step consists in updating the resources usage by the scheduled tasks (line~\ref{alg_desc:alg_schedule_task_update_resources}).

Algorithm~\ref{alg:scheduleTask} has a time complexity of $O(n^2)$, where $n = \sum_k n_k$. Searching for a feasible starting time $s^k_i$ (line~\ref{alg_desc:alg_schedule_task_get_sk}) requires at most $O(n-1)$ time if task $J^k_i$ has $n-1$ predecessors. Searching for the event $e$ such that $t(e) = s^k_i$ (line~\ref{alg_desc:alg_schedule_task_get_previous_event}) requires at most $O(\log |\mathcal{EL}|)$ since a balanced tree data structure is used for the event list.

In the worst case, for a set $\mathcal{P}$ of plans with cardinality $|\mathcal{P}| = K$, the maximum number of events in $\mathcal{EL}$ is $2n$ if we suppose that all the tasks are scheduled consecutively with a minimal lag between each task, that is

$$|\mathcal{S}(e)|=1 \wedge |\mathcal{C}(e)| = 0~\vee $$
$$|\mathcal{S}(e)|=0 \wedge |\mathcal{C}(e)| = 1, ~\forall e \in \mathcal{EL}.$$ 

The while loop (lines~\ref{alg_desc:alg_schedule_task_loop_until_end}-\ref{alg_desc:alg_schedule_task_end_while}) iterates over the event list $\mathcal{EL}$. Since there are less than $2n$ events, the while loop takes $O(n)$ time. 

Searching for the event following $f$ in $\mathcal{EL}$ (line~\ref{alg_desc:alg_schedule_task_assign_nextf_to_g}) takes at most $O(\log |\mathcal{EL}|)$ time. Checking if the temporal constraints are satisfied at $t(e)$ (line~\ref{alg_desc:alg_schedule_task_check_constraints}) takes $O(1)$ time. The remaining operations inside the loop require constant time.

Removing all the tasks $J^k_i \in \Pi_k$ from each event $e \in \mathcal{EL}$ (line~\ref{alg_desc:alg_schedule_task_remove_plan_from_events}) takes $O(n_k \log |\mathcal{EL}|)$.

Adding the task $J^k_i$ to the schedule $S_w$ (line~\ref{alg_desc:alg_schedule_task_add_to_solution}) requires at most $O(\log n)$ since an efficient balanced tree data structure is used for the schedules. Adding the task $J^k_i$ to the set $\mathcal{S}_e$ (line~\ref{alg_desc:alg_schedule_task_add_to_S}) requires constant time.

Adding the event $e$ to the event list $\mathcal{EL}$ (line~\ref{alg_desc:alg_schedule_task_add_e_to_EL}) requires $O(\log |\mathcal{EL}|)$ time. 

 The event list update takes $O(\log |\mathcal{EL}|)$ if the event $g$ has to be inserted in $\mathcal{EL}$ (lines~\ref{alg_desc:alg_schedule_task_insert_g_1},\ref{alg_desc:alg_schedule_task_insert_g_2}), otherwise the insertion of $J^k_i$ into $\mathcal{C}(f)$ (line~\ref{alg_desc:alg_schedule_task_insert_task_in_Cf}) requires $O(1)$ time.

		\begin{algorithm}[!h]
			\caption{ScheduleTask}
			\label{alg:scheduleTask}
			\begin{algorithmic}[1]
				\REQUIRE A task $J^k_i$, a schedule $S_w$, an event list $\mathcal{EL}$
				\ENSURE $true$ if $J^k_i$ has been scheduled in $S_w$, $false$ otherwise.
				\STATE $s^k_i = max(W_s, r^k_i, ~max_{j\in pred^k_i}(C^k_j + \delta^k_{ij}))$     \label{alg_desc:alg_schedule_task_get_sk}
				\STATE $e \leftarrow $ GetEvent($s^k_i$, $\mathcal{EL}$) \label{alg_desc:alg_schedule_task_get_previous_event}
				\STATE $\mu \leftarrow p^k_i; f \leftarrow  e$  \label{alg_desc:alg_schedule_task_get_mu_and_f}
				\WHILE{($f$ is not the last event of $\mathcal{EL}) \wedge (\mu > 0)$}  \label{alg_desc:alg_schedule_task_loop_until_end}
				\STATE $g \leftarrow next_f$  \label{alg_desc:alg_schedule_task_assign_nextf_to_g}
				\IF{$(b_{\rho}(f) = 0,~\forall \rho \in \mathcal{R}^k_i) \wedge $\\$($CheckConstraints$(t(e),J^k_i))$}  \label{alg_desc:alg_schedule_task_check_constraints}
				\STATE $\mu \leftarrow max(0,\mu-t(g)+t(f))$
				\STATE $f\leftarrow g$
				\ELSE
				\STATE $\mu \leftarrow p^k_i$; $e \leftarrow g$; $f \leftarrow g$
				\ENDIF
				\ENDWHILE \label{alg_desc:alg_schedule_task_end_while}
				    
				\IF{\NOT CheckConstraints$(t(e),J^k_i)$}  \label{alg_desc:alg_schedule_task_check_final_constraints}
				\STATE remove all tasks $J^k_i \in \Pi_k$ from $e$, $\forall e \in \mathcal{EL}$ \label{alg_desc:alg_schedule_task_remove_plan_from_events}
				\STATE mark $\Pi_k$ as unschedulable \label{alg_desc:alg_schedule_task_mark_plan_as_unschedulable}
				\RETURN \textit{false}
				\ENDIF
				    
				\STATE Add($J^k_i$, $S_w$)  \label{alg_desc:alg_schedule_task_add_to_solution}
				\STATE Add($J^k_i$, $\mathcal{S}$($e$)) \label{alg_desc:alg_schedule_task_add_to_S}
				\IF{$e \notin \mathcal{EL}$}  \label{alg_desc:alg_schedule_task_check_e_in_el}
				    \STATE $\mathcal{EL} \leftarrow \mathcal{EL} \cup e$  \label{alg_desc:alg_schedule_task_add_e_to_EL}
				\ENDIF
				
				\IF{$t(e) + p^k_i = t(f)$}  \label{alg_desc:events_update_start}
				\STATE Add($J^k_i$, $\mathcal{C}$($f$)) \label{alg_desc:alg_schedule_task_insert_task_in_Cf}
				\ELSIF{$t(e) + p^k_i > t(f)$}
				\STATE insert $g=(t(e)+p^k_i, \emptyset, \{t\}, (b_{\rho}(f))_{\forall \rho \in \mathcal{R}^k_i} )$ in $\mathcal{EL}$  \label{alg_desc:alg_schedule_task_insert_g_1}
				\STATE $f \leftarrow g$
				\ELSE
				\STATE insert $g=(t(e)+p^k_i, \emptyset, \{t\}, (b_{\rho}(pred_f))_{\forall \rho \in \mathcal{R}^k_i} )$ in $\mathcal{EL}$  \label{alg_desc:alg_schedule_task_insert_g_2}
				\STATE $f \leftarrow g$
				\ENDIF  \label{alg_desc:events_update_end}
				 
				\STATE $b_{\rho}(g) =1~\forall \rho \in \mathcal{R}^k_i,~\forall g$ between $e$ and $pred_f$ in $\mathcal{EL}$ \label{alg_desc:alg_schedule_task_update_resources}   
				\RETURN \textit{true}
			\end{algorithmic}
		\end{algorithm}

Algorithm~\ref{alg:schedulePlanWithSamePriority} ($SchedulePlanSet$) schedules a set $\mathcal{P}_{\alpha_k}$ of plans such that each plan $\Pi_k \in \mathcal{P}_{\alpha_k}$ has a priority value $\alpha_k$.
The idea of the algorithm is to find, at each iteration, the plan who minimizes the sum of the size of the idle time windows generated by its tasks. The algorithm starts by creating a copy $V$ of the plan set $\mathcal{P}_{\alpha_k}$ (line~\ref{alg_schedule_multi:init_v}). The while loop (lines \ref{alg_schedule_multi:plans_schedule_loop_start}-\ref{alg_schedule_multi:plans_schedule_loop_end}) iterates until there is no plan left in $V$ to schedule. At each iteration, two variables $minIdleTime$ and $bestPlan$ are used to keep the next candidate plan to schedule. The loop at (\ref{alg_schedule_multi:single_plan_test_loop_start}-\ref{alg_schedule_multi:single_plan_test_loop_end}) iterates over the plans in $V$ and searches for the candidate plan to schedule. Each plan $\Pi_k$ is scheduled into a temporary schedule $S_{temp}$ (line~\ref{alg_schedule_multi:schedule_pik_into_temp_solution}). At each iteration, the algorithm chooses the plan $\Pi_k \in \mathcal{P}_{\alpha_k}$ that minimizes the sum of the time differences between $t(e)$ and $t(pred_e)$ (line~\ref{alg_schedule_multi:calc_idle_time_sum}), for each event $e$ such that $J^k_i \in \mathcal{S}(e)$, $\forall J^k_i \in \Pi_k$.
		
Algorithm~\ref{alg:schedulePlanWithSamePriority} takes at most $O(K^2n^3)$ time. The main loop (lines~\ref{alg_schedule_multi:plans_schedule_loop_start}-\ref{alg_schedule_multi:plans_schedule_loop_end}) is repeated until there is some plan left to schedule, and it is repeated at most $O(K)$ time. The inner loop (lines~\ref{alg_schedule_multi:single_plan_test_loop_start}-\ref{alg_schedule_multi:single_plan_test_loop_end}) schedule each plan $\Pi_k \in \mathcal{P}_{\alpha_k}$ and keeps the plan which minimize the idle times between each task and its predecessor. The inner loop takes $O(Kn^3)$ time.

		\begin{algorithm}[!h]
			\caption{SchedulePlanSet}
			\label{alg:schedulePlanWithSamePriority}
			\begin{algorithmic}[1]
				\REQUIRE A set $\mathcal{P}_{\alpha_k} \subset \mathcal{P}$ of plans, a schedule $S_w$, an event list $\mathcal{EL}$
				\ENSURE A set $\mathcal{U}$ of unscheduled plans
				\STATE $V \leftarrow \mathcal{P}_{\alpha_k}$  \label{alg_schedule_multi:init_v}
				\STATE $\mathcal{U} \leftarrow \emptyset$

				\WHILE{$V \neq \emptyset$}  \label{alg_schedule_multi:plans_schedule_loop_start}
				
        \STATE $minIdleTime \leftarrow +\infty$
        \STATE $bestPlan \leftarrow \emptyset$        
				
				\FORALL{$\Pi_k \in \mathcal{P}_{\alpha_k}$}   \label{alg_schedule_multi:single_plan_test_loop_start}
				\STATE $S_{temp} \leftarrow S_w$
				\STATE $\mathcal{EL}_{temp} \leftarrow \mathcal{EL}$
				\STATE $success \leftarrow schedulePlan(\Pi_k, S_{temp}, \mathcal{EL}_{temp})$ \label{alg_schedule_multi:schedule_pik_into_temp_solution}
				\IF{$success$}
				  \STATE $it = \sum_{i} (t(e)-t(pred_{e})), \forall J^k_i \in \mathcal{S}(e)$ \label{alg_schedule_multi:calc_idle_time_sum}
				  \IF{$it \leq minIdleTime$}
				    \STATE $minIdleTime \leftarrow it$
				    \STATE $bestPlan \leftarrow \Pi_k$
				  \ENDIF
				\ELSE
  				\STATE $Remove(\Pi_{k}, V)$  \label{alg_schedule_multi:remove_best_plan_from_v}
				\ENDIF
				\ENDFOR \label{alg_schedule_multi:single_plan_test_loop_end}
				\STATE $success \leftarrow  schedulePlan(bestPlan, S_w, \mathcal{EL})$  \label{alg_schedule_multi:schedule_best_plan}
				\IF{\NOT $success$}
				\STATE $Add(bestPlan, \mathcal{U})$
				\ENDIF
				\STATE $Remove(bestPlan, V)$  \label{alg_schedule_multi:remove_best_plan_from_v}
				
				\ENDWHILE       \label{alg_schedule_multi:plans_schedule_loop_end}
				
				\RETURN $\mathcal{U}$
			\end{algorithmic}
		\end{algorithm}
		
		Algorithm~\ref{alg:check_temporal_constraints} ($CheckConstraints$) checks if a starting time $t(e)$ satisfies the temporal constraints, discussed in~\ref{Temporal constraints}, for a task $J^k_i$. If the temporal constraints are satisfied, the algorithm returns \textit{true}.	
		
Algorithm~\ref{alg:check_temporal_constraints} requires constant time.
		
		\begin{algorithm}[!h]
			\caption{CheckConstraints}
			\label{alg:check_temporal_constraints}
			\begin{algorithmic}[1]
				\REQUIRE A task $J^k_i$, a starting time $t(e)$ for $J^k_i$
				\ENSURE $true$ if $J^k_i$ satisfies the temporal constraints, $false$ otherwise.
				\IF {$\left(t(e) \in [r^k_i,d^k_i]\right) \wedge \left(t(e) \in [W_s,W_e]\right) \wedge \left(t(e)+p^k_i \in [W_s,W_e]\right)$}  \label{alg_desc:check_temporal_constraint}
				\RETURN \textit{true}
				\ENDIF
				\RETURN \textit{false}
			\end{algorithmic}
		\end{algorithm}

		In Table~\ref{table:table_of_functions} are resumed the functions employed by the proposed method, together with a short description and their time complexity.
		
		\begin{table}
			\begin{center}
				\begin{tabulary}{.50\textwidth}{L{24em}C{5em}}
					\hline
					\textbf{Function name and description}   & \textbf{Time complexity} \\ \hline
					$BuildSchedule(\mathcal{P})$ \\
					 Entry point algorithm. It is responsible \newline for assembling a \textit{feasible} schedule for a set $\mathcal{P}$ of plans. & $O(K^3n^3)$ \\ \hline
					$SchedulePlan(\Pi_k, S_w, \mathcal{EL})$ \\
					 Schedule the plan $\Pi_k$ into the schedule $S_w$. & $O(n^3)$ \\ \hline
					$SchedulePlanSet(\mathcal{P}_{\alpha_k},S_w, \mathcal{EL})$  \\ 
					 Schedule a set of plans $\mathcal{P}_{\alpha_k}$ that have the same priority value into the schedule $S_w$. & $O(K^2n^3)$ \\ \hline
					$ScheduleTask(J^k_i,S_w, \mathcal{EL})$& \\
					  Schedule the task $J^k_i$ into the schedule $S_w$. & $O(n^2)$  \\ \hline
					$SortPlans(\mathcal{P})$  \\
					 Sort a set of plans $\mathcal{P}$ according to their precedence relations by using $TopologicalSort$ algorithm. Then, the plans that belong to the same frontier in the topological sorting are sorted by their priority value. & $O(K\log K + n_k)$ \\ \hline
					$TopologicalSort(\mathcal{P})$  \\
					 Sort a set of plans $\mathcal{P}$ according to their precedence relations by using a topological sorting algorithm. It returns a sorted copy of the plan set given as input. & $O(K+n_k)$ \\ \hline
					$GetEvent(s^k_i, \mathcal{EL})$  \\
					 Get the event $e \in \mathcal{EL}$ such that $t(e) = s^k_i$. If such event does not exists, a new event $e$ is created at $t(e) = s^k_i$. & $O(\log |\mathcal{EL}|)$ \\ \hline
					$Get(\mathcal{P})$ Get the highest priority plan from $\mathcal{P}$. & $O(\log K)$ \\ \hline
				\end{tabulary}  
			\end{center}
			\caption{Descriptions of the functions used by the scheduling algorithm.}
			\label{table:table_of_functions}
		\end{table}

\begin{exmp}
\label{example_2}
In this example we show how different plans are scheduled according to the proposed heuristic. We start from a feasible schedule containing two plans $\Pi_1$ and $\Pi_2$, and we try to schedule in order the plans $\Pi_3$, $\Pi_4$ and $\Pi_5$. In Table~\ref{table:example_2_table} are listed the parameters of the plans of tasks $\mathcal{P}=\{\Pi_1,\Pi_2,\Pi_3,\Pi_4,\Pi_5\}$.

\begin{table}[!h]
\centering
\begin{tabular}{lllllll}
\hline
        & $r^k_i$ & $d^k_i$ & $p^k_i$  & $\delta^k_{ij}$ & $\mathcal{R}^k_i$ & $J^k_i \prec J^k_j$\\ \hline
$J^1_1$ & $1$ & $7$  & $4$  & $0$ & $\{1\}$ & $J^1_1 \prec J^1_2$\\ \hline
$J^1_2$ & $5$ & $8$  & $2$  & $1$ & $\{3\}$ & \\ \hline
$J^2_1$ & $4$ & $7$  & $2$  & $0$ & $\{2\}$ & $J^2_1 \prec J^2_2$\\ \hline
$J^2_2$ & $5$ & $9$  & $3$  & $0$ & $\{1\}$ & \\ \hline
$J^3_1$ & $1$ & $8$  & $3$  & $0$ & $\{3\}$ & \\ \hline
$J^4_1$ & $2$ & $7$  & $2$  & $0$ & $\{2\}$ & $J^4_1 \prec J^4_2$\\ \hline
$J^4_2$ & $3$ & $7$  & $1$  & $2$ & $\{3\}$ & \\ \hline
$J^5_1$ & $5$ & $10$  & $3$  & $0$ & $\{2\}$ & $J^5_1 \prec J^5_2$\\ \hline
$J^5_2$ & $5$ & $11$  & $1$  & $0$ & $\{1,3\}$ & \\ \hline
\end{tabular}
\caption{The parameters of the tasks of the plans in Example~\ref{example_2}.}
\label{table:example_2_table}
\end{table}

Let $\Pi_3$ be the next plan to be scheduled, which has only one task $J^3_1$. 
The earliest starting time $s^3_1$ is set to the release time $r^3_1 = 1$. The event $e_1 \in \mathcal{EL}$ has a time instant $t(e) = r^3_1 = 1$, thus $e_1$ is the first event in which $J^3_1$ can be scheduled. The resource constraints in $e_1$ are satisfied for $J^3_1$, so it can be scheduled at $t(e_1) = 2$. The resource usages $b_3(e_1)$ and $b_3(e_2)$ are finally updated. 

Table~\ref{tbl:example_insertion_plan_3} shows the resulting Gantt diagram and the event list $\mathcal{EL}$ after the insertion of the plan $\Pi_3$.

\begin{table*}[htb]
  \centering
  \begin{minipage}[l]{.2\textwidth}
    \begin{tabular}{lllllll}
				\hline
				$e$  & $t(e)$ & $\mathcal{S}(e)$                    & $\mathcal{C}(e)$          & $b_{1}(e)$ & $b_{2}(e)$ & $b_{3}(e)$ \\ \hline
				$e1$ & $2$    & $\{J^{1}_{1},\bm{\textcolor{red}{J^{3}_{1}}}\}$      & $\emptyset$               & $1$        & $0$        & \bm{\textcolor{red}{$1$}}         \\ \hline
				$e2$ & $4$    & $\{J^{2}_{1}\}$                     & $\emptyset$               & $1$        & $1$        & \bm{\textcolor{red}{$1$}}        \\ \hline
				$e3$ & $5$    & $\emptyset$           			        & $\{\bm{\textcolor{red}{J^{3}_{1}}}\}$ 			& $1$        & $1$        & $0$        \\ \hline
				$e4$ & $6$    & $\{J^{2}_{2}\}$           & $\{J^{2}_{1},J^{1}_{1}\}$ & $1$        & $0$        & $0$        \\ \hline
			  $e5$ & $7$    & $\{J^{1}_{2}\}$           & $\emptyset$ & $1$        & $0$        & $1$        \\ \hline
				$e6$ & $9$    & $\emptyset$                         & $\{J^{1}_{2},J^{2}_{2}\}$           & $0$        & $0$        & $0$        \\ \hline
			\end{tabular}
  \end{minipage}\hfill
  \begin{minipage}[r]{0.45\textwidth}
    \includegraphics[width=\textwidth]{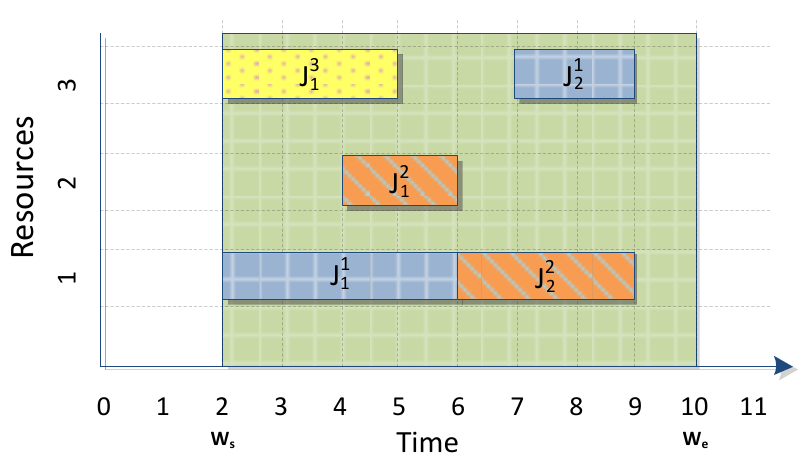}
  \end{minipage}
  \caption{Insertion of the plan $\Pi_3$ into a feasible schedule.}
  \label{tbl:example_insertion_plan_3}
\end{table*}

Following, plan $\Pi_4$ is the next plan to be scheduled. The earliest starting time $s^4_1$ for $J^4_1$ is $s^4_1 = t(e_1) = W_s$. At $e_1$ the temporal and resource constraints for $J^4_1$ are satisfied, so it is scheduled at $s^4_1 = t(e_1) =2$, and then the resource usages $b_2(e_1)$ is updated. Following, an insertion point is searched for the task $J^4_2$. The earliest starting time for $J^4_2$ is $s^4_2 = t(e_4)=6$, since there is a time lag between $J^4_2$ and its predecessor $J^4_1$. 
The event $e_4$ is a good insertion point for the task $J^4_2$, since here the precedence, resource and temporal constraints are met. $J^4_2$ is finally scheduled at $s^4_2= t(e_4) = 6$. The resource usage $b_3(e_4)$ is updated. Since a feasible insertion point has been found for both $J^4_1$ and $J^4_2$, the plan $\Pi_4$ is scheduled correctly and the schedule generated is considered as feasible schedule. 

Table~\ref{tbl:example_insertion_plan_4} shows the resulting Gantt diagram and the event list $\mathcal{EL}$ after the insertion of the plan $\Pi_4$.

\begin{table*}[htb]
  \centering
  \begin{minipage}[l]{.2\textwidth}
    \begin{tabular}{lllllll}
				\hline
				$e$  & $t(e)$ & $\mathcal{S}(e)$                              & $\mathcal{C}(e)$          & $b_{1}(e)$ & $b_{2}(e)$ & $b_{3}(e)$ \\ \hline
				$e1$ & $2$    & $\{J^{1}_{1},J^{3}_{1},\bm{\textcolor{red}{J^{4}_{1}}}\}$      & $\emptyset$               & $1$        & \bm{\textcolor{red}{$1$}}         & $1$        \\ \hline
				$e2$ & $4$    & $\{J^{2}_{1}\}$                               & $\{\bm{\textcolor{red}{J^{4}_{1}}}\}$               & $1$        & $1$        & $1$        \\ \hline
				$e3$ & $5$    & $\emptyset$           			          & $\{J^{3}_{1}\}$ 			& $1$        & $1$        &  $0$        \\ \hline
				$e4$ & $6$    & $\{\bm{\textcolor{red}{J^{4}_{2}}},J^{2}_{2}\}$                     & $\{J^{2}_{1},J^{1}_{1}\}$ & $1$        & $0$        & \bm{\textcolor{red}{$1$}}        \\ \hline
			  $e5$ & $7$    & $\{J^{1}_{2}\}$           & $\{\bm{\textcolor{red}{J^{4}_{2}}}\}$ & $1$        & $0$        & $1$        \\ \hline
				$e6$ & $9$    & $\emptyset$                                   & $\{J^{1}_{2}, J^{2}_{2}\}$           & $0$        & $0$        & $0$        \\ \hline
			\end{tabular}
  \end{minipage}\hfill
  \begin{minipage}[r]{0.45\textwidth}
    \includegraphics[width=\textwidth]{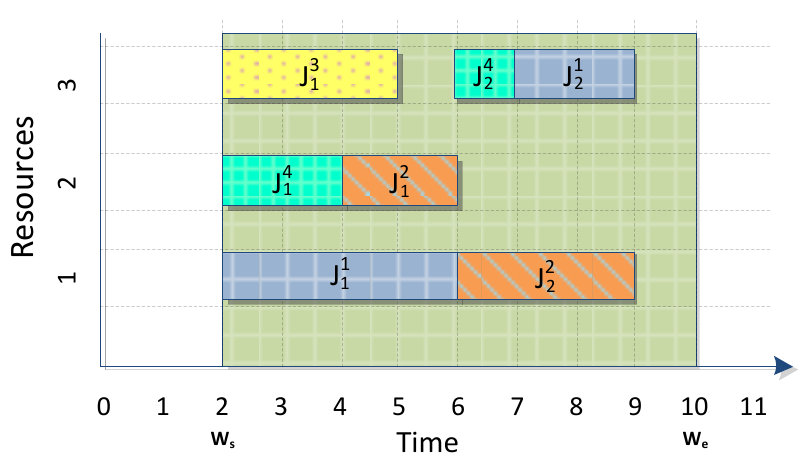}
  \end{minipage}
  \caption{Insertion of the plan $\Pi_4$ into a feasible schedule.}
  \label{tbl:example_insertion_plan_4}
\end{table*}

Finally, plan $\Pi_5$ is the last plan to be scheduled. The earliest starting time for $J^5_1$ is $s^5_1 = 5$. At $t(e_3)=5$ the temporal and resource constraints for $J^5_1$ are not satisfied ($b_2(e_3)=1$). The next event $e_4$ is a good insertion point for the task $J^5_1$, since here the resource constraint and also the temporal constraints are met. The task $J^5_1$ is finally scheduled at \mbox{$s^5_1= t(e_4)$}. Following, an insertion point is searched for the task $J^5_2$, which consumes both the resources $1$ and $3$. The earliest starting time for $J^5_2$ is $s^5_2=5$. The resources constraints are not satisfied in the events $e_3$, $e_4$ and $e_5$. The event $e_6$ is the first feasible insertion point for the task $J^5_2$, since at $t(e_6)$ the precedence and temporal constraints are met. Also, in $e_6$ both resources $1$ and $3$ are available ($b_1(e_6)=0$ and $b_3(e_6)=0$). The task $J^5_2$ is finally scheduled at $s^5_2= t(e_6)$, and the usages of resources $1$ and $3$, respectively $b_1(e_6)$ and $b_3(e_6)$, are updated. Since a feasible insertion point has been found for both $J^5_1$ and $J^5_2$, the plan $\Pi_5$ is scheduled correctly and the schedule generated is considered as feasible schedule. 

Table~\ref{tbl:example_insertion_plan_5} shows the resulting Gantt diagram and the event list $\mathcal{EL}$ after the insertion of the plan $\Pi_5$.

\begin{table*}[htb]
  \centering
  \begin{minipage}[l]{.2\textwidth}
    \begin{tabular}{lllllll}
				\hline
				$e$  & $t(e)$ & $\mathcal{S}(e)$                    & $\mathcal{C}(e)$          & $b_{1}(e)$ & $b_{2}(e)$ & $b_{3}(e)$ \\ \hline
				$e1$ & $2$    & $\{J^{1}_{1},J^{3}_{1},J^{4}_{1}\}$      & $\emptyset$               & $1$        & $1$        & $1$        \\ \hline
				$e2$ & $4$    & $\{J^{2}_{1}\}$                     & $\{J^{4}_{1}\}$               & $1$        & $1$        & $1$        \\ \hline
				$e3$ & $5$    & $\emptyset$           			        & $\{J^{3}_{1}\}$ 			& $1$        & $1$        & $0$        \\ \hline
				$e4$ & $6$    & $\{J^{4}_{2},J^{2}_{2},\bm{\textcolor{red}{J^{5}_{1}}}\}$           & $\{J^{2}_{1},J^{1}_{1}\}$ & $1$        & \bm{\textcolor{red}{$1$}}        & $1$        \\ \hline
			  $e5$ & $7$    & $\{J^{1}_{2}\}$           & $\{J^{4}_{2}\}$ & $1$        & \bm{\textcolor{red}{$1$}}        & $1$        \\ \hline				
				$e6$ & $9$    & $\{\bm{\textcolor{red}{J^{5}_{2}}}\}$                         & $\{J^{1}_{2},J^{2}_{2},\bm{\textcolor{red}{J^{5}_{1}}}\}$           & \bm{\textcolor{red}{$1$}}        & $0$        & \bm{\textcolor{red}{$1$}}        \\ \hline
				$e7$ & $10$   & $\emptyset$                         & $\{\bm{\textcolor{red}{J^{5}_{2}}}\}$               & $0$        & $0$        & $0$        \\ \hline
			\end{tabular}
  \end{minipage}\hfill
  \begin{minipage}[r]{0.45\textwidth}
    \includegraphics[width=\textwidth]{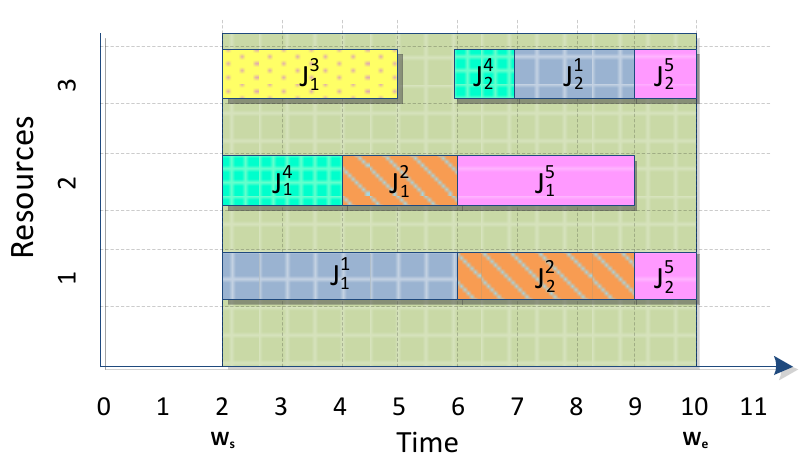}
  \end{minipage}
  \caption{Insertion of the plan $\Pi_5$ into a feasible schedule.}
  \label{tbl:example_insertion_plan_5}
\end{table*}

\end{exmp}

\section{Computational results}
In this section we present the computational results for the proposed heuristic. All the experiments were run on an entry-level machine (Intel i5-4210U, 6GB RAM, Linux OS). The algorithms were entirely written in Java (version $8$).

We run different experiments of our algorithm by using a benchmark that depicts a realist scenario of a mission carried out by a Remote Piloted Air Vehicle (RPAS), which set of plans, listed in Table~\ref{table:benchmark_plans}, has to be scheduled. The validity of the mission is constrained between $0$ and $180$ seconds. Therefore, the time window $[W_s,W_e]$ is fixed to $[0,180]$.

\begin{table*}
\centering
\begin{tabular}{cccccccccccc}
\hline
$\Pi_k$ & $n_k$ & $\alpha_k$ & $\mathcal{R}^k$ & $[r^k,d^k]$ & $p^k_1$ & $p^k_2$ & $p^k_3$ & $p^k_4$ & $\delta^k_{ij}$ & $\Pi_k \prec \Pi_j$ & $J^k_i \prec J^k_j$ \\ \hline 
$\Pi_{1}$ & $2$ & $3$ & $\{1, 2\}$ & $[0,180]$ & $3$ & $3$ &  &  &  &  & $J^{1}_{1} \prec J^{1}_{2}$ \\ \hline 
$\Pi_{2}$ & $2$ & $3$ & $\{3,4\}$ & $[0,180]$ & $4$ & $4$ &  &  &  &  &  \\ \hline 
$\Pi_{3}$ & $1$ & $3$ & $\{5\}$ & $[0,180]$ & $3$ &  &  &  &  &  &  \\ \hline 
$\Pi_{4}$ & $1$ & $3$ & $\{6\}$ & $[0,180]$ & $3$ &  &  &  &  &  &  \\ \hline 
$\Pi_{5}$ & $1$ & $3$ & $\{7\}$ & $[0,180]$ & $1$ &  &  &  &  &  &  \\ \hline 
$\Pi_{6}$ & $1$ & $3$ & $\{8\}$ & $[0,180]$ & $1$ &  &  &  &  &  &  \\ \hline 
$\Pi_{7}$ & $4$ & $1$ & $\{5,9,10,11\}$ & $[80,160]$ & $20$ & $20$ & $20$ & $20$ &  &  & $J^{7}_{3} \prec J^{7}_{4}$ \\ \hline 
$\Pi_{8}$ & $4$ & $1$ & $\{5,9,10,11\}$ & $[40,120]$ & $20$ & $20$ & $20$ & $20$ &  &  & $J^{8}_{3} \prec J^{8}_{4}$ \\ \hline 
$\Pi_{9}$ & $4$ & $1$ & $\{5,9,10,11\}$ & $[40,120]$ & $20$ & $20$ & $20$ & $20$ &  &  & $J^{9}_{3} \prec J^{9}_{4}$ \\ \hline 
$\Pi_{10}$ & $3$ & $5$ & $\{1, 2, 10\}$ & $[50,80]$ & $5$ & $5$ & $5$ &  &  &  & $J^{10}_{1} \prec J^{10}_{2}$ \\ \hline 
$\Pi_{11}$ & $4$ & $5$ & $\{5,9,10,11\}$ & $[80,120]$ & $20$ & $20$ & $20$ & $20$ &  & $\Pi_{10} \prec \Pi_{11}$ & $J^{11}_{3} \prec J^{11}_{4}$ \\ \hline 
$\Pi_{12}$ & $4$ & $1$ & $\{5,9,10,11\}$ & $[0,80]$ & $20$ & $20$ & $20$ & $20$ &  &  & $J^{12}_{3} \prec J^{12}_{4}$ \\ \hline 
$\Pi_{13}$ & $3$ & $5$ & $\{1, 2, 10\}$ & $[0,40]$ & $5$ & $5$ & $5$ &  &  &  & $J^{13}_{1} \prec J^{13}_{2}$ \\ \hline 
$\Pi_{14}$ & $4$ & $5$ & $\{5,9,10,11\}$ & $[40,80]$ & $20$ & $20$ & $20$ & $20$ &  & $\Pi_{13} \prec \Pi_{14}$ & $J^{14}_{3} \prec J^{14}_{4}$ \\ \hline 
$\Pi_{15}$ & $3$ & $6$ & $\{1, 2, 10\}$ & $[0,20]$ & $5$ & $5$ & $5$ &  &  &  & $J^{15}_{1} \prec J^{15}_{2}$ \\ \hline 
$\Pi_{16}$ & $4$ & $6$ & $\{5,9,10,11\}$ & $[20,80]$ & $20$ & $20$ & $20$ & $20$ &  & $\Pi_{15} \prec \Pi_{16}$ & $J^{16}_{3} \prec J^{16}_{4}$ \\ \hline 
$\Pi_{17}$ & $2$ & $8$ & $\{3,4\}$ & $[60,120]$ & $2$ & $2$ &  &  &  &  & $J^{17}_{1} \prec J^{17}_{2}$ \\ \hline 
$\Pi_{18}$ & $2$ & $8$ & $\{3,4\}$ & $[60,120]$ & $2$ & $2$ &  &  &  &  & $J^{18}_{1} \prec J^{18}_{2}$ \\ \hline 
$\Pi_{19}$ & $2$ & $8$ & $\{3,4\}$ & $[60,120]$ & $2$ & $2$ &  &  &  &  & $J^{19}_{1} \prec J^{19}_{2}$ \\ \hline 
$\Pi_{20}$ & $3$ & $6$ & $\{1, 2, 10\}$ & $[0,20]$ & $5$ & $5$ & $5$ &  &  &  & $J^{20}_{1} \prec J^{20}_{2}$ \\ \hline 
$\Pi_{21}$ & $4$ & $6$ & $\{6,9,10,11\}$ & $[20,60]$ & $20$ & $20$ & $20$ & $20$ &  & $\Pi_{20} \prec \Pi_{21}$ & $J^{21}_{3} \prec J^{21}_{4}$ \\ \hline 
$\Pi_{22}$ & $2$ & $6$ & $\{14,10\}$ & $[30,70]$ & $2$ & $2$ &  &  &  &  &  \\ \hline 
$\Pi_{23}$ & $4$ & $1$ & $\{6,9,10,11\}$ & $[40,90]$ & $20$ & $20$ & $20$ & $20$ &  &  & $J^{23}_{3} \prec J^{23}_{4}$ \\ \hline 
$\Pi_{24}$ & $4$ & $1$ & $\{6,9,10,11\}$ & $[80,150]$ & $20$ & $20$ & $20$ & $20$ &  &  & $J^{24}_{3} \prec J^{24}_{4}$ \\ \hline 
$\Pi_{25}$ & $4$ & $1$ & $\{6,9,10,11\}$ & $[80,130]$ & $20$ & $20$ & $20$ & $20$ &  &  & $J^{25}_{3} \prec J^{25}_{4}$ \\ \hline 
$\Pi_{26}$ & $4$ & $1$ & $\{6,9,10,11\}$ & $[130,160]$ & $20$ & $20$ & $20$ & $20$ &  &  & $J^{26}_{3} \prec J^{26}_{4}$ \\ \hline 
$\Pi_{27}$ & $2$ & $4$ & $\{14,10\}$ & $[0,180]$ & $2$ & $2$ &  &  &  &  &  \\ \hline 
$\Pi_{28}$ & $3$ & $6$ & $\{1, 2, 10\}$ & $[80,120]$ & $5$ & $5$ & $5$ &  &  &  & $J^{28}_{1} \prec J^{28}_{2}$ \\ \hline 
$\Pi_{29}$ & $2$ & $6$ & $\{3,4\}$ & $[120,140]$ & $2$ & $2$ &  &  &  & $\Pi_{28} \prec \Pi_{29}$  & $J^{29}_{1} \prec J^{29}_{2}$ \\ \hline 
$\Pi_{30}$ & $2$ & $6$ & $\{14,10\}$ & $[120,140]$ & $2$ & $2$ &  &  &  &  &  \\ \hline 
$\Pi_{31}$ & $4$ & $1$ & $\{6,9,10,11\}$ & $[150,190]$ & $20$ & $20$ & $20$ & $20$ &  &  & $J^{31}_{3} \prec J^{31}_{4}$ \\ \hline 
$\Pi_{32}$ & $2$ & $4$ & $\{3,4\}$ & $[120,140]$ & $2$ & $2$ &  &  &  &  & $J^{32}_{1} \prec J^{32}_{2}$ \\ \hline 
\end{tabular}
\caption{The benchmark used for evaluating the performance of the proposed method.}
\label{table:benchmark_plans}
\end{table*}

We first evaluated the proposed heuristic on the benchmark listed in Table~\ref{table:benchmark_plans}. Then, we evaluated the performance of the proposed heuristic on different benchmarks obtained by modifying the set of plans listed in Table~\ref{table:benchmark_plans}. These modified benchmarks are opportunely set up in such a way that the constraints are more or less relaxed. Thus, we studied eight different scenarios, which results are listed in Table~\ref{table:experiments_duration}.
Following are listed the differences between each scenario:

\begin{itemize}
  \item \textit{Scenario 1}: set of plans listed in Table~\ref{table:benchmark_plans};
  \item \textit{Scenario 2}: in this scenario all the plans have the same temporal window $[r^k,d^k]$ fixed to $[0,180]$, and the plans are all tied by precedence relations, so that \mbox{$\forall i,j, j>i, \Pi_i \prec \Pi_j$};
  \item \textit{Scenario 3}: in this scenario we modified the release time of each plan in such way to have the maximum number of events;
  \item \textit{Scenario 4}: in this scenario we duplicated all the plans with priority $6$ and $8$;
  \item \textit{Scenario 5}: in this scenario we duplicated all the plans with priority $1$ and $3$;
  \item \textit{Scenario 6}: in this scenario we used the same benchmark as the scenario 1, but we used a time window $[W_s,W_e]$ fixed to $[0,90]$;
  \item \textit{Scenario 7}: in this scenario we used the same benchmark as the scenario 1, but we used a time window $[W_s,W_e]$ fixed to $[0,270]$;
  \item \textit{Scenario 8}: in this scenario we used the same benchmark as the scenario 1, but each plan has a duplicate.
\end{itemize}

Table~\ref{table:experiments_duration} presents the execution times for each scenario. Each execution time is calculated as the mean of the execution times of ten runs of the algorithm on the same benchmark.

\begin{table}[!h]
\centering
\begin{tabular}{c|ccccc}
           & $K=|\mathcal{P}|$ & $Kn_k$ & $|\mathcal{P}_s|$ & $|\mathcal{P}_s|n_k$ & Time (in ms)   \\ \hline
Scenario 1 & $32$ & $91$ & $24$  & $56$ & $\sim183$                         \\ \hline
Scenario 2 & $32$ & $91$ & $24$ & $67$ & $\sim183$                         \\ \hline
Scenario 3 & $32$ & $91$ & $24$ & $64$ & $\sim219$                         \\ \hline
Scenario 4 & $42$ & $118$ & $30$ & $68$ & $\sim316$                         \\ \hline
Scenario 5 & $47$ & $135$ & $29$ & $64$ & $\sim345$                         \\ \hline
Scenario 6 & $32$ & $91$ & $16$ & $35$ & $\sim197$                         \\ \hline
Scenario 7 & $32$ & $91$ & $24$ & $60$ & $\sim228$                         \\ \hline
Scenario 8 & $64$ & $182$ & $37$ & $81$ & $\sim538$                         \\ \hline
\end{tabular}
\caption{Execution times of the eight scenarios.}
\label{table:experiments_duration}
\end{table}

The obtained computational results of the algorithm gave us a first proof of concept concerning the complexity of the scheduling problem we address. Moreover, the results are yet satisfactory in the context of scheduling plans for RPAS. As the problem of scheduling plans of tasks is difficult due to its tight constraints, we conclude that the quality of the schedules provided by the proposed heuristic is good with respect to the complexity of the evaluated benchmarks.

\section{Future works}
\subsection{Variable resource availability}
The proposed heuristic considers only resources with a fixed availability of $B_\rho = 1,~\forall \rho \in \mathcal{R}$. A generalization of the algorithm could take into account resources with variable availability values, so that one resource could be used by different tasks at the same time. Augmenting the resource availability could also lead to better schedules in terms of number of plans scheduled into a fixed time window, as showed in Example~\ref{example_resource_usage}.

\begin{exmp}
\label{example_resource_usage}
Let $\mathcal{P}=\{\Pi_1,\Pi_2\}$ be a set of two plans $\Pi_1$ and $\Pi_2$ which have each one task, respectively $J^1_1$ ($r^1_1 = 1$, $d^1_1 = 6$, $p^1_1 = 3$) and $J^2_1$ ($r^2_1 = 4$, $d^2_1 = 10$, $p^2_1 = 4$). In Figure~\ref{fig:varying_res_capacity} are showed two possible schedules for the task $J^2_1$. If the resource $1$ has a maximum availability $B_1=1$ (Figure~\ref{fig:task_schedule_with_res_capacity_1}), the earliest starting time for the task $J^2_1$ would be $6$, because otherwise the usage of resource would exceed its availability. Instead, if the availability of the resource is $2$ (Figure~\ref{fig:task_schedule_with_res_capacity_2}), the earliest starting time would be set to $4$ (that is, the release time of the task), and so the use of the resource is maximized.
Also, in the schedule showed in Figure~\ref{fig:task_schedule_with_res_capacity_2} the time interval $[5,8]$ can be employed for scheduling other tasks, whereas this is not possible in the schedule in Figure~\ref{fig:task_schedule_with_res_capacity_1}.
		
		\begin{figure}[!h]
			\centering
			\begin{subfigure}{.90\linewidth}
				\centering
				\includegraphics[width=\linewidth]{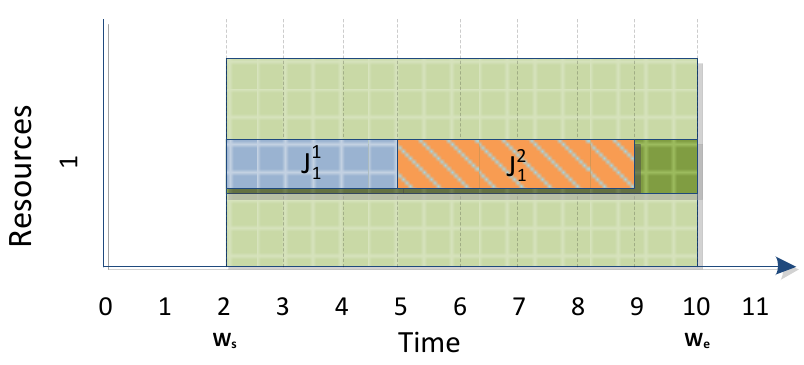}
				\caption{}
				\label{fig:task_schedule_with_res_capacity_1}
			\end{subfigure}%
			
			\hfill
			
			\begin{subfigure}{.90\linewidth}
				\centering
				\includegraphics[width=\linewidth]{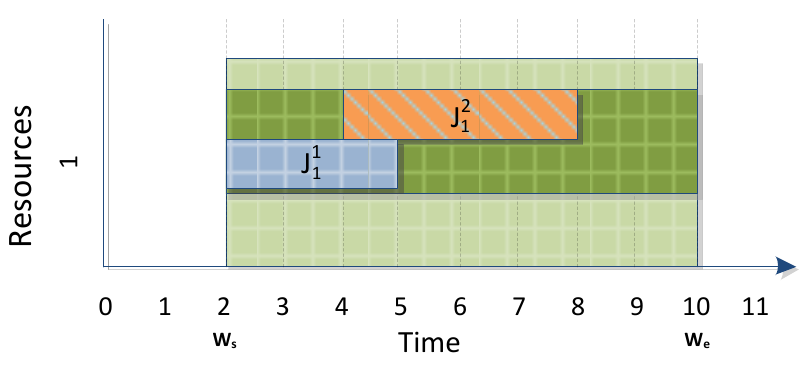}
				\caption{}
				\label{fig:task_schedule_with_res_capacity_2}
			\end{subfigure}%
			\caption{Scheduling of task $J^2_1$, which has a release time of $4$, into a resource with availability $B_1=1$ (\ref{fig:task_schedule_with_res_capacity_1}) and a resource with availability $B_1=2$ (\ref{fig:task_schedule_with_res_capacity_2}).}
			\label{fig:varying_res_capacity}
		\end{figure}

\end{exmp}

\subsection{$p^k_i$ as function of time}
The value of the processing time $p^k_i$ for a task $J^k_i$ could be modeled as a function of time $p^{k}_{i}(t)$ in such a way that the processing time of a task $J^k_i$ can vary according to the time instant in which it is scheduled. To motivate this extension, we present in Example~\ref{example_rpas_function_time} a case study of a RPAS that has to accomplish some in-flight operation. 

\begin{exmp}
\label{example_rpas_function_time}
Let us consider a RPAS that has to accomplish a mission where it has to follow a straight trajectory and collect data from an object $R$ located in the terrain during a fixed time range, as depicted in Figure~\ref{fig:rpas}. During the the time range in which the RPAS is within the object's range it has to send information to the object $R$ in the terrain. The operation of sending information could require more time when the RPAS is far from the object, but less time when it is near the object. Thus, the processing time function related to the operation of sending information could be modeled as an inverse Gaussian function, as showed in Figure~\ref{fig:inv_gaussian}, where $d_{max}$ and $d_{min}$ are respectively the maximum and the minimum distance between the RPAS and the object $R$.

\begin{figure}[!h]
	\centering
	\includegraphics[scale=0.85]{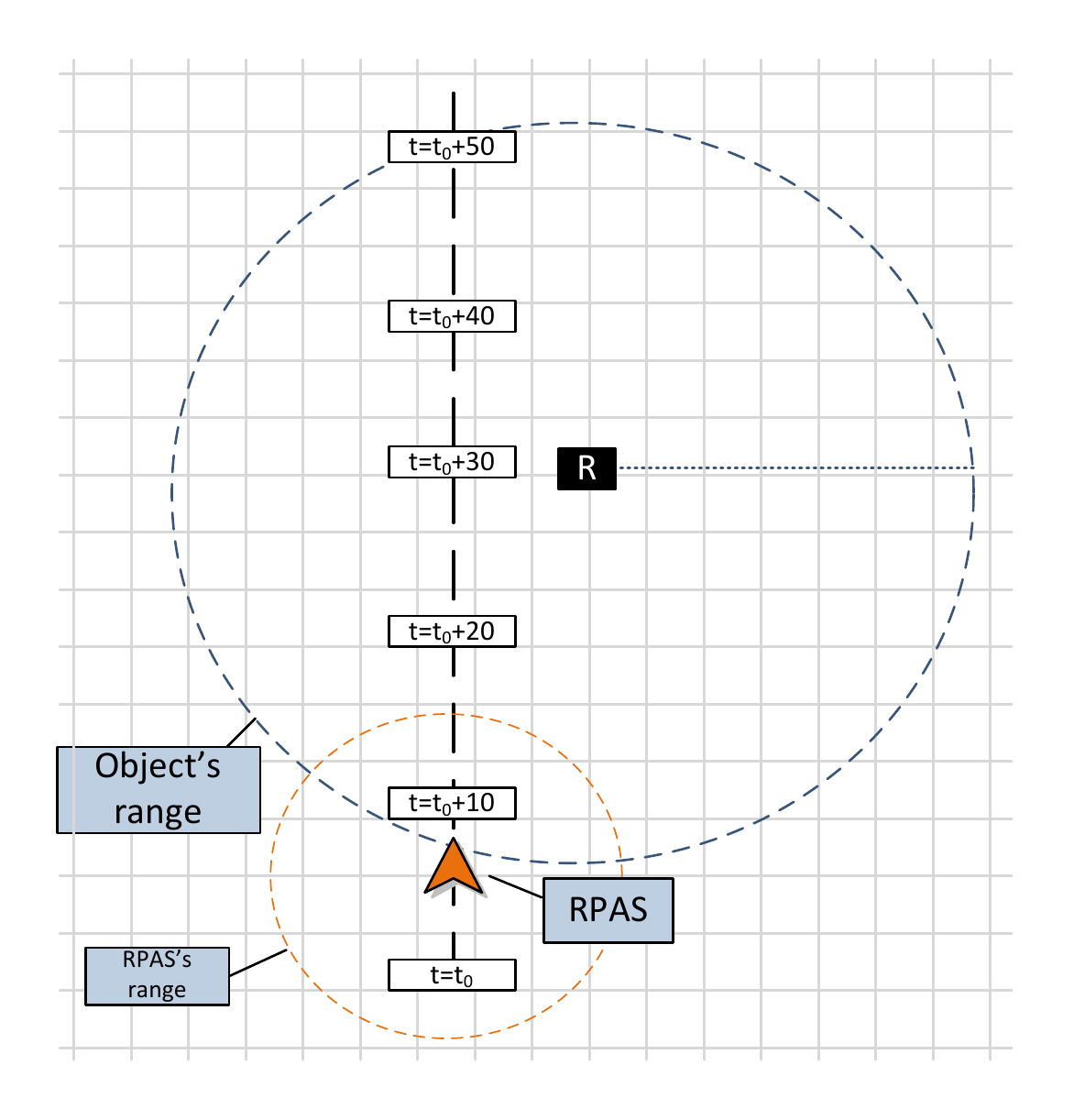}
	\caption{An example of RPAS that has to communicate informations to an object $R$. The processing time of the tasks communicating with $R$ could vary according to the distance between the object $R$ and the RPAS with respect to the time instant $t$.}
	\label{fig:rpas}
\end{figure}

\begin{figure}[!ht]
	\centering
	\includegraphics[scale=0.7]{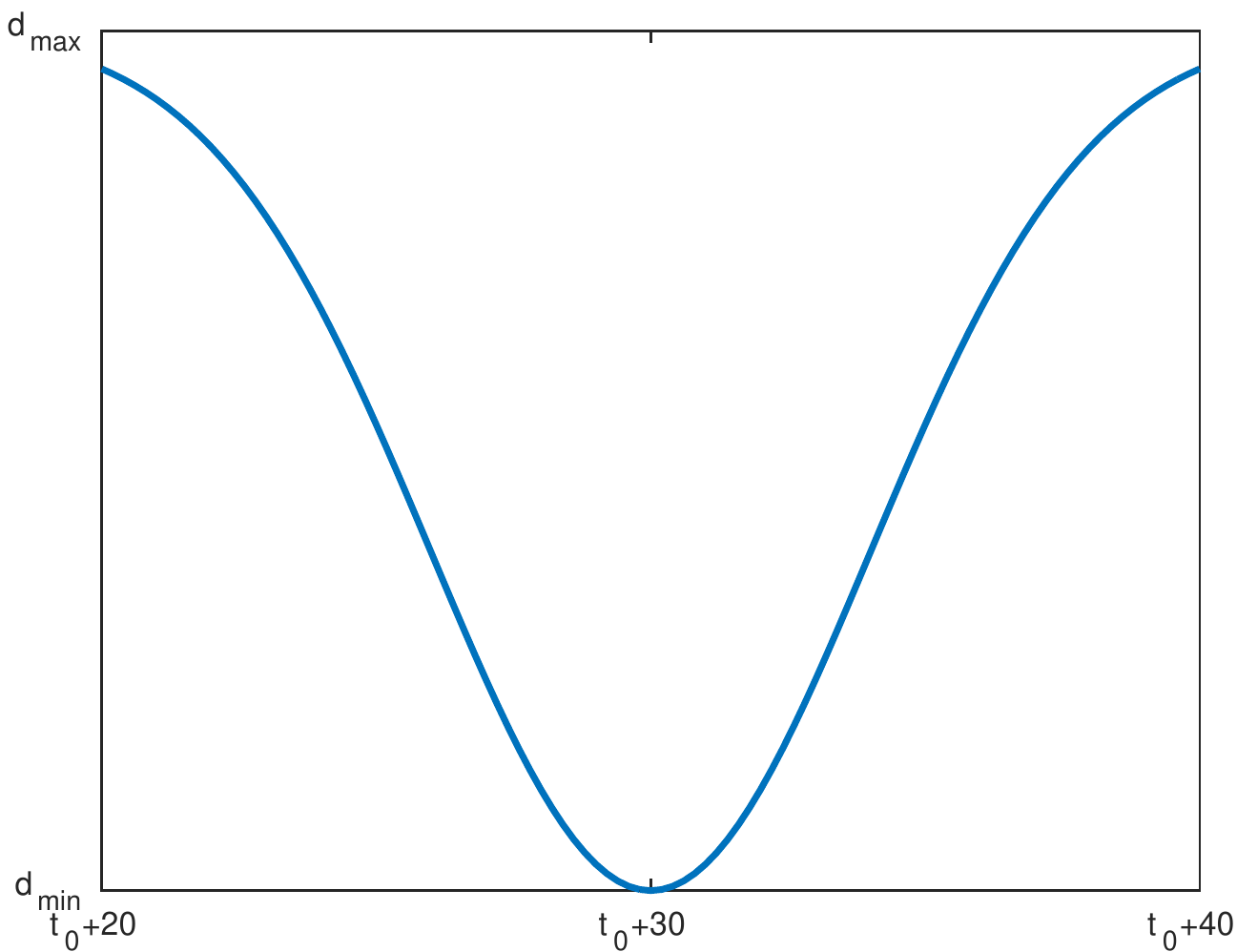}
	\caption{An inverse Gaussian used to model the processing time $p(t)$ of the the operation of sending information to a fixed object for the RPAS in Figure~\ref{fig:rpas}.}
	\label{fig:inv_gaussian}
\end{figure}
		
\end{exmp}

\subsection{Alternative schedules}
\begin{figure}[htbp]
			\centering
			\includegraphics[scale=0.75]{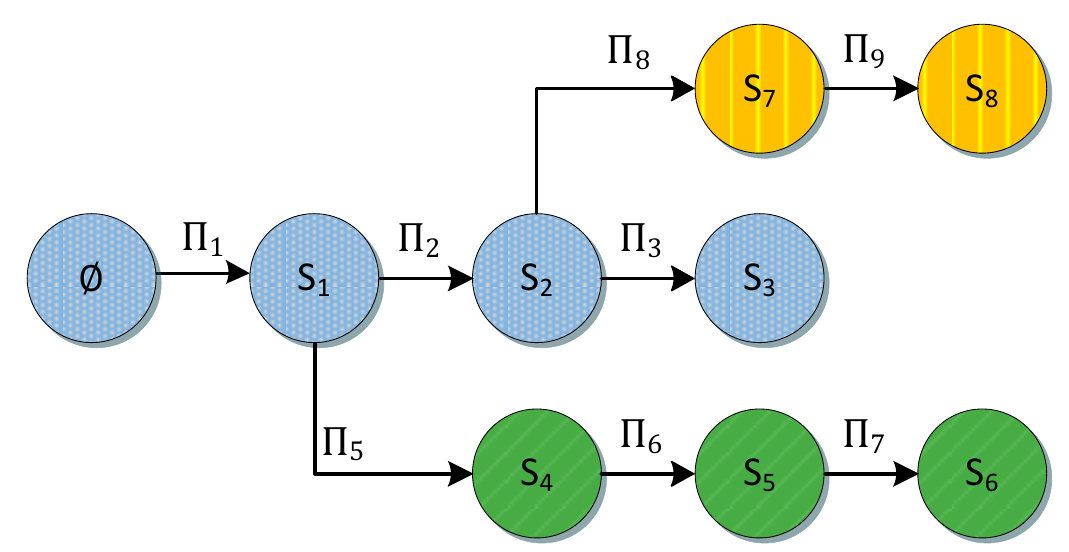}
			\caption{The graph representation of a feasible schedule with four plans (in blue), and two possible alternative schedule (yellow and green).}
			\label{fig:graph_solution_representation}
		\end{figure}

		The proposed heuristic works in an incremental fashion, adding plans in order to a working schedule $S_w$ and keeping a set $\mathcal{P}_s$ of successfully scheduled plans, as well as a set $\mathcal{P}_f$ of unscheduled plans. The incremental process of construction of a feasible schedule can be represented as an directed acyclic graph, as shown in the example graph in Figure~\ref{fig:graph_solution_representation}. According to this representation, each vertex represents a feasible schedule, while a directed arc $(i,j)$, which has label $\Pi_k$, represents the insertion of the plan $\Pi_k$ into the schedule $S_i$, that leads to a new feasible schedule $S_j$.
		
The incremental manner of constructing the schedules has the advantage of being able to return a feasible schedule at each iteration of the algorithm.

Since the proposed method builds only a graph with no branches (no alternative schedule paths), a possible extension could consider different metrics to build alternative paths in the graph of solutions. For example, as the proposed heuristic returns a set $\mathcal{P}_f$ of plans unscheduled due to constraints violations, one possible extension could increase the priority values of the unscheduled plans in $\mathcal{P}_f$ while lowering those of a subset of feasible plans in $\mathcal{P}_s$. In this manner, alternative paths in the solution graph could be found. Also, these alternative paths could maximize further the objective function, thus allowing the heuristic method to escape from the local optima.

\section{Conclusions}

This article presented a heuristic algorithm for the problem of scheduling plans of tasks. The algorithm may be employed in realistic scenarios where the classical scheduling technique can not be used due to the considered constraints.

	
\end{document}